\pdfoutput=1
\documentclass[11pt]{article}

\usepackage[final]{acl}

\usepackage{times}
\usepackage{latexsym}

\usepackage[T1]{fontenc}
\usepackage[utf8]{inputenc}

\usepackage{microtype}

\usepackage{inconsolata}

\usepackage{graphicx}

\usepackage{makecell} % For multiline cell tables

\usepackage{booktabs}
\usepackage{amsmath}
\usepackage{caption} % for side to side graphs 

\newcommand{\laleaderboard}{\mbox{\textsc{La Leaderboard} }}

\definecolor{darkblue}{rgb}{0,0,0.44}

\title{La Leaderboard: A Large Language Model Leaderboard for Spanish Varieties and Languages of Spain and Latin America}

\author{
  \textbf{María Grandury\textsuperscript{1,2}}, % maria.grandury@somosnlp.org
  \textbf{Javier Aula-Blasco\textsuperscript{3}}, % javier.aulablasco@bsc.es
  \textbf{Júlia Falcão\textsuperscript{3}}, % julia.falcao@bsc.es
  \textbf{Clémentine Fourrier\textsuperscript{4}}, % clementine@huggingface.co
\\
  \textbf{Miguel González\textsuperscript{2}}, % miguel.gonsaiz@upm.es
  \textbf{Gonzalo Martínez\textsuperscript{5}}, % gonzmart@pa.uc3m.es
  \textbf{Gonzalo Santamaría\textsuperscript{6}} % gonzalo.santamaria@iic.uam.es 
\\%[1ex]
  \textbf{Rodrigo Agerri\textsuperscript{7}},
  \textbf{Nuria Aldama\textsuperscript{6}}, % nuria.aldama@iic.uam.es
  \textbf{Luis Chiruzzo\textsuperscript{13}}, % luis.chiruzzo@gmail.com 
  \textbf{Javier Conde\textsuperscript{2}}, % javier.conde.diaz@upm.es
  \textbf{Helena Gómez\textsuperscript{12}}, % helena.gomez@iimas.unam.mx 
  \textbf{Marta} \\ \textbf{Guerrero\textsuperscript{6}}, % marta.guerrero@iic.uam.es
  \textbf{Guido Ivetta\textsuperscript{11}}, % guidoivetta@mi.unc.edu.ar 
  \textbf{Natalia López\textsuperscript{6}}, % natalia.lopez@iic.uam.es 
  \textbf{Flor Miriam Plaza-del-Arco\textsuperscript{9}}, % flor.plaza@unibocconi.it
  \textbf{María Teresa} \\ \textbf{Martín-Valdivia\textsuperscript{10}}, % maite@ujaen.es
  \textbf{Helena Montoro\textsuperscript{6}}, % helena.montoro@iic.uam.es
  \textbf{Carmen Muñoz\textsuperscript{6}}, % carmen.munoz@iic.uam.es
  \textbf{Pedro Reviriego\textsuperscript{2}}, % pedro.reviriego@upm.es 
  \textbf{Leire Rosado\textsuperscript{6}}, % leire.rosado@iic.uam.es
  \\
  \textbf{Alejandro Vaca\textsuperscript{8}}, % alejandro_vaca0@hotmail.com
  \textbf{María Estrella Vallecillo-Rodríguez\textsuperscript{10}}, % mevallec@ujaen.es
  \textbf{Jorge Vallego\textsuperscript{}}, % jorge.vallego@theh4rmonyproject.org
  \textbf{Irune Zubiaga\textsuperscript{7}} %
\\
\\ % 1º orden de main contributors, 2º número de datasets donados
  \textsuperscript{1}SomosNLP,
  \textsuperscript{2}ETSIT, Universidad Politécnica de Madrid, % ETSI de Telecomunicación
  \textsuperscript{3}Barcelona Supercomputing Center, \\ % (BSC-CNS), \\
  \textsuperscript{4}Hugging Face,
  \textsuperscript{5}Universidad Carlos III de Madrid, 
  \textsuperscript{6}Instituto de Ingeniería del Conocimiento, \\
  \textsuperscript{7}Centro HiTZ - Ixa, Universidad del País Vasco UPV/EHU,
  \textsuperscript{8}LenguajeNatural.AI, \\
  \textsuperscript{9}LIACS, Leiden University,
  \textsuperscript{10}Universidad de Jaén,
  \textsuperscript{11}Universidad Nacional de Córdoba, \\
  \textsuperscript{12}Universidad Nacional Autónoma de México,
  \textsuperscript{13}Universidad de la República, Uruguay
\\
  \small{
    \textbf{Correspondence:} \href{mailto:maria.grandury@somosnlp.org}{maria.grandury@somosnlp.org}
  }
}

\begin{document}
\maketitle
\begin{abstract}

%Leaderboards are the most widespread and convenient way to measure the performance evolution of Large Language Models (LLMs).
Leaderboards showcase the current capabilities and limitations of Large Language Models (LLMs).
To motivate the development of LLMs that represent the linguistic and cultural diversity of the Spanish-speaking community, we present \laleaderboard\footnote{\url{https://hf.co/spaces/la-leaderboard/la-leaderboard}}, the first open-source leaderboard to evaluate generative LLMs in languages and language varieties of Spain and Latin America.
\laleaderboard is a community-driven project that aims to establish an evaluation standard for everyone interested in developing LLMs for the Spanish-speaking community. This initial version combines 66 datasets in Basque, Catalan, Galician, and different Spanish varieties, showcasing the evaluation results of 50 models.
To encourage community-driven development of leaderboards in other languages, we explain our methodology, including guidance on selecting the most suitable evaluation setup for each downstream task.
In particular, we provide a rationale for using fewer few-shot examples than typically found in the literature, aiming to reduce environmental impact and facilitate access to reproducible results for a broader research community.
% In particular, we argue that, contrary to common practice, using less than 5 examples in few-shot settings allows for a fair evaluation for both base and instructed models. In fact, this may have the added benefit of helping to reduce the environmental impact and facilitate access to reproducible results for a broader research community.
\end{abstract}

% Our story:
%- we believe leaderboards guide what we optimize LLMs for since they show the capabilities and limitations of current LLMs but there was no leaderboard for the spanish speaking community
%- we have created one collaborating with many people, here's how so you can do it too
%- since we've created the leaderboard we have evaluated models and here is what we've found about the current sota of LLMs in our languages

\section{Introduction}
\label{sec:introduction}

The evaluation of multilingual Large Language Models (LLMs) is challenging. LLMs are expected to perform a large variety of tasks, from problem-solving to text summarization, all in multiple languages \cite{LLM_evaluation_survey}.
In this context, leaderboards have emerged as one of the standard approaches for evaluating and comparing LLMs in a transparent manner. As we cannot improve what we cannot measure, it is important to develop leaderboards that enable a more comprehensive evaluation of LLMs across linguistic boundaries, contributing to the development of culturally aware AI systems that can serve diverse global linguistic communities.

\begin{figure}[t]
  \includegraphics[width=\columnwidth]{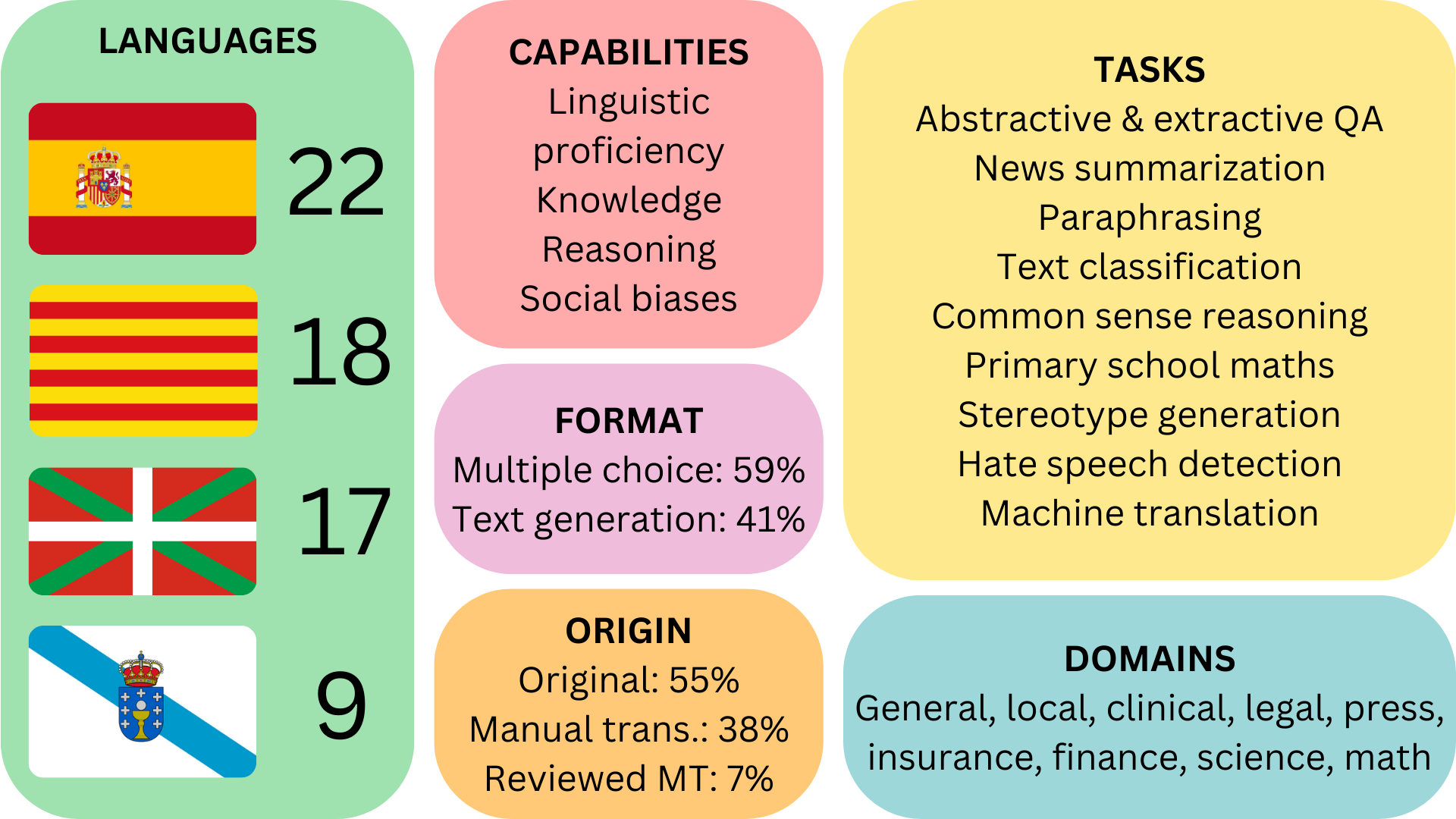}
  \caption{Summary of the evaluation datasets included in \laleaderboard.
  Disclaimer: A country does not represent a language; flags are used for simplicity.}
  \label{fig:task_summary}
\end{figure}

Spanish is one of the most spoken languages worldwide, with more than 600 million speakers \cite{fernandez2024demografia}. It is the predominant language in 21 countries,
%, which results in different language varieties. 
where it coexists with other languages. 
Many people use Spanish and the local language in their daily activities.
Spain has four official languages: Spanish, Catalan, Basque, and Galician. While Catalan and Galician are Romance languages closely related to Spanish, Basque is one of the world's few language isolates \cite{Campbell2010LanguageIA}.
In Latin America (LATAM), there are hundreds of indigenous languages, such as Guarani and Náhuatl,
%Quechua, and Aymara. These languages have diverse origins and, in some cases, 
which have influenced local Spanish varieties \cite{lustig1996mba}.
From a sociolinguistic point of view, this creates a unique scenario for multilingual LLM evaluation.
Moreover, knowing which LLMs perform best in these languages can have deep implications for multilingual communication \cite{strassel-tracey-2016-lorelei}.
% https://www.sciencedaily.com/releases/2010/01/100127144549.htm
% OPTIONAL añadir otra cita referente a la comunidad hispanohablante
% these languages -> low-resource languages? local languages?

Existing leaderboards predominantly focus on English or a small set of high-resource languages \cite{open-llm-leaderboard-v2, mialon2023gaia, open_medical_llm_leaderboard, 2023opencompasschinesebench}. 
While Spanish is often included in multilingual leaderboards, evaluation datasets are typically limited and translated, either by machines
%\footnote{\url{https://hf.co/spaces/openGPT-X/european-llm-leaderboard}}
\cite{OcciglotEuroLeaderboard}, failing to capture the linguistic richness of the language \cite{lostintranslation} or by humans\footnote{\url{https://hf.co/datasets/openai/MMMLU}}, still failing to represent the target culture \cite{singh2024globalmmluunderstandingaddressing}
Moreover, despite the growing presence of LLMs in multilingual settings, no leaderboard currently evaluates a combination of languages spoken in Spain and Latin America.
This lack of representation limits the development of models that can truly serve these communities \cite{mager2018challenges}.

To address this gap, we introduce \laleaderboard,
%\footnote{\url{https://hf.co/spaces/la-leaderboard/la-leaderboard}}, % already linked in first page
the first open-source leaderboard designed to evaluate generative LLMs based on the needs of the Spanish-speaking community.
Beyond the initial set of languages that includes Spanish and the official languages of Spain (Basque, Catalan, and Galician), \laleaderboard is designed to evolve, gradually expanding to encompass more languages and linguistic varieties, ensuring it reflects the rich diversity of the global community.
This new leaderboard consists of a diverse set of evaluation tasks (see Figure~\ref{fig:task_summary}) designed in a way that reflects the nuances and actual usage of the target languages. It is a community-driven initiative
launched by the \#Somos600M Project \cite{grandury2024somos600m},
aiming to foster the development of LLMs that better represent the linguistic and cultural diversity of the Spanish-speaking world. We share our approach to inspire other linguistic communities to create similar leaderboards.

The main contributions of this work are:
\begin{itemize}
    \item We present the community-based methodology used to create the first open-source leaderboard for evaluating generative LLMs in Spanish and the official languages of Spain, with a scalable framework designed to include more languages and language varieties over time.
    \item We introduce a logical and resource-efficient approach to few-shot configurations, enabling accessible and reproducible evaluations for the wider community.
    \item We provide a comprehensive analysis of state-of-the-art (SOTA) LLMs, providing insights into their strengths and limitations in Spanish, Catalan, Basque, and Galician.
\end{itemize}

By addressing the linguistic and cultural
diversity of Spain and LATAM, \laleaderboard sets a new standard for multilingual LLM evaluation, which encourages the development of models that are not only linguistically competent but also culturally aware. 
%ultimately driving progress in Natural Language Processing for the benefit of our whole community.

\section{Related Work}
\label{sec:related_work}

\paragraph{Benchmarks} Several benchmarks have been developed to evaluate the performance of LLMs in tasks like language understanding \cite{wang2019superglue}, general knowledge \cite{MMLU}, reasoning \cite{sakaguchi2019winograndeadversarialwinogradschema}, or mathematical problem solving \cite{Mathmeasuring}. There are also efforts to develop holistic benchmarks or evaluation suites that provide a comprehensive evaluation of different capabilities of LLMs \cite{liang2023holistichelm, eval-harness, lighteval, open-llm-leaderboard-v2, bigbench}.
% Commented out after rebuttal to leave place for more detailed comparison with ODESIA and CLUB.
% However, as LLM performance improves continuously, benchmarks soon become saturated \cite{kiela2021dynabenchrethinkingbenchmarkingnlp} or contaminated \cite{xu2024benchmark}, which makes it necessary to continuously develop new tasks. For example, MMLU-Pro \cite{mmlu-pro} has been proposed to replace MMLU, and shortly after that, MMLU-Pro+ was proposed as an additional improvement \cite{mmlu-pro+}, with even harder tests being developed \cite{phan2025humanitylastexam}. 
% https://www.evidentlyai.com/llm-guide/llm-benchmarks#common-llm-benchmarks

\paragraph{Multilingual and multicultural benchmarks} LLMs are now trained in multiple high-resource languages at the same time
%\footnote{\url{https://hf.co/collections/BSC-LT/salamandra-66fc171485944df79469043a}} 
\cite{ali2024teuken7bbaseteuken7binstructeuropean, martins2024eurollmmultilinguallanguagemodels, qwen25, mistral7b}, which means that the benchmarks must reflect this linguistic diversity.
A common approach is machine translating English tests \cite{multilingualevaluation, openai2023gpt4}. However, translation errors may add noise to the results, making them less reliable \cite{lostintranslation}. Furthermore, each language has its nuances, preferred styles, and cultural background, which unrevised machine translation may fail to capture \cite{plaza-del-arco-etal-2020-emoevent, singh2024globalmmluunderstandingaddressing}.
Ideally, specific test sets should be originally written in the target language or manually adapted \cite{nangia2020crows} to capture the richness and cultural and linguistic subtleties associated with it. This is slowly becoming the trend as shown by the recently released language-specific \cite{ItalianTests, MedFrenchTest} and multilingual culturally aware \cite{romanou2024include, salazar2025kaleidoscopeinlanguageexamsmassively, myung2025blendbenchmarkllmseveryday, romero2024cvqaculturallydiversemultilingualvisual} benchmarks.

% Leaderboard Collection: https://hf.co/collections/clefourrier/leaderboards-and-benchmarks-64f99d2e11e92ca5568a7cce
\paragraph{Leaderboards} Benchmarks are the pieces of the LLM evaluation puzzle that provide valuable but fragmented information on their performance. Leaderboards and arenas use these evaluation sets to compare the performance of LLMs in a neutral, third-party manner through automatic evaluations \cite{mialon2023gaia} or human judgments \cite{chatbotarena}. On some community-oriented leaderboards \cite{open-llm-leaderboard-v2}, anyone can submit their LLMs for evaluation, and the tools, tests, and results are open, allowing for reproducibility. This represents a good way to drive progress in LLM development by enabling people with limited compute to compare their models to the current SOTA.
% There is literature alerting about issues that leaderboards may have: https://aclanthology.org/2020.emnlp-main.393/ or https://aclanthology.org/2024.acl-long.744/

\paragraph{Multilingual leaderboards} Leaderboards exhibit the same shortcomings as benchmarks when evaluating languages other than English. To address this problem, specific leaderboards are being developed in different languages such as Italian \cite{ItalianTests},
%Korean \cite{park2024openKorean, kim2024openKorean2},
Korean \cite{kim2024openKorean2},
Chinese \cite{2023opencompasschinesebench}, Arabic \cite{open_arabic_llm_leaderboard} or Polish \cite{jassem2025llmzszl_Polish}.

\paragraph{Spanish leaderboards} Focusing on the Spanish language, the ODESIA leaderboard\footnote{\url{https://leaderboard.odesia.uned.es}} by UNED NLP features 14 bilingual Spanish-English datasets from discriminative shared tasks. The team evaluates 10 LLMs. While submissions are open, the evaluation datasets are private, avoiding task contamination \cite{salido2025bilingualevaluationlanguagemodels} but making it impossible to reproduce the results locally.
Regarding text generation, Spanish is represented in the Chatbot Arena\footnote{\url{https://lmarena.ai}}, which features a dedicated category, and in SCALE’s private leaderboard\footnote{\url{https://scale.com/leaderboard/spanish}}. However, both exclusively evaluate a fixed set of models.
The only existing leaderboard including a language from Spain or Latin America other than Spanish is CLUB\footnote{\url{https://club.aina.bsc.es}}, developed by the BSC as part of the AINA Project, which combines 8 Catalan datasets and evaluates 5 BERT and RoBERTa-based models. \\ % Añadir "\\", queda mejor separado pero añade 2 líneas

In this work, we present the methodology used to create a comprehensive, fully open-source leaderboard for languages and language varieties from Spain and Latin America that assesses different capabilities of generative models, including domain knowledge, information extraction, linguistic proficiency, and ethical aspects. \laleaderboard aims to serve as a reference for the Spanish-speaking scientific community, fostering the development of more robust and culturally adequate LLMs.

\section{\laleaderboard}
\label{sec:laleaderboard}

\laleaderboard is a community-driven initiative that brings together 66 datasets in Spanish, Catalan, Basque, and Galician, covering diverse tasks and domains. Public since September 23, 2024, \laleaderboard has received over 15,000 visits in five months and currently showcases evaluation results from 50 models.

% Methodology paper (here's how to choose relevant datasets/leverage community work/build a leaderboard/...)

\subsection{Data Collection}
\label{sec:data_collection}
Most of the datasets in \laleaderboard were donated by 13 research groups. Initially, these contributions were received through a publicly shared Google Form (Appendix~\ref{sec:appendix_data_collection}) or direct outreach. In particular, 7 datasets were specifically created for \laleaderboard (AQuAS, ClinTreatES, ClinDiagnosES, HumorQA, SpaLawEx, TELEIA, and RAGQuAS). We also included widely used open-source benchmarks such as Belebele.
% Unclear whether people started contacting you or you them. I would add a small sentence on how the project reached out to contributors, then the fact that this generated 3 avenues: standard datasets, datasets obtained through outreach, datasets provided by the community/labs.

\laleaderboard keeps expanding with dataset contributions such as CONAN-EUS and VeritasQA. These new connections are bidirectional: we actively share this initiative in relevant conferences and reach out to research groups, while others contact us upon discovering \laleaderboard.
Beyond collecting existing datasets, we are also fostering collaborations to enhance the representation of languages and linguistic varieties across Latin America.

% Responsible guidelines: discuss whether and how consent was obtained from people whose data you’re using/curating
To thank research groups for their donations, we include in \laleaderboard's interface the corresponding logo and dataset citation. Moreover, the dataset authors are acknowledged in this paper. % and those who made valuable contributions to the project were invited as co-authors.
% Ethics review: the contributors who helped to construct the benchmark should be co-authors

% FUTURE WORK: Quality control, analyze in depth the kind of errors models do to see if they come from the model or the dataset (Clementine)

\subsection{Task Construction}
\label{sec:task_construction}

\subsubsection{Datasets}
\label{sec:datasets}

Including diverse evaluation datasets is essential for building a comprehensive leaderboard. This section discusses the key axes that guided their selection.
Table~\ref{tab:tasks_per_language} enumerates the datasets organized by language and task type, while Table~\ref{tab:upcoming_datasets} shows the upcoming datasets that have been recently donated and not yet evaluated. In Appendix~\ref{sec:appendix_datasets}, we provide the citations and further details about the datasets, including origin and domain.

% STYLE: En lugar de parrafo, puedes usar \begin{tabularx}{\textwidth}{lX} que se ajusta directamente al texto cada columna. (Flor)

\begin{table*}[!ht]
%\small
\centering
\scalebox{0.95}{
\begin{tabular}{p{2.5cm}p{2.8cm}p{2.8cm}p{2.8cm}p{2.8cm}}
\toprule
\makecell[c]{\textbf{Task Type}}  & 
\makecell[c]{\textbf{Spanish}}  & 
\makecell[c]{\textbf{Catalan}}  & 
\makecell[c]{\textbf{Basque}}  & 
\makecell[c]{\textbf{Galician}} \\
\hline % con \midrule no cabe la tabla
\makecell[l]{Common-sense \\ reasoning} & 
\makecell[l]{copa\_es \\ xstorycloze\_es} & 
\makecell[l]{copa\_ca \\ xstorycloze\_ca} & 
\makecell[l]{xcopa\_eu \\ xstorycloze\_eu} & 
\makecell[l]{xstorycloze\_gl} \\
\hline
\makecell[l]{Ethics} & 
\makecell[l]{crows\_pairs\_es} & 
\makecell[l]{crows\_pairs\_ca} & 
\makecell[l]{--} & 
\makecell[l]{--} \\
\hline
\makecell[l]{Linguistic \\ acceptability} & 
\makecell[l]{escola} & 
\makecell[l]{catcola} & 
\makecell[l]{--} & 
\makecell[l]{galcola} \\
\hline
\makecell[l]{Math} & 
\makecell[l]{mgsm\_direct\_es} & 
\makecell[l]{mgsm\_direct\_ca} & 
\makecell[l]{mgsm\_direct\_eu} & 
\makecell[l]{mgsm\_direct\_gl} \\
\hline
\makecell[l]{NLI} & 
\makecell[l]{wnli\_es \\ xnli\_es} & 
\makecell[l]{teca \\ wnli\_ca \\ xnli\_ca} & 
\makecell[l]{epec\_koref\_bin
\\ qnli\_eu,  wiceu \\ wnli\_eu,  xnli\_eu} & 
\makecell[l]{xnli\_gl}
\\
\hline
\makecell[l]{Paraphrasing} & 
\makecell[l]{paws\_es \\ parafrases\_sushi} & % humorqa
\makecell[l]{parafraseja \\ paws\_ca} &
\makecell[l]{--} & 
\makecell[l]{parafrases\_gl \\ paws\_gl} \\
\hline
\makecell[l]{Question \\ answering} & 
\makecell[l]{aquas \\ clindiagnoses \\ clintreates \\ spalawex \\ teleia \\ ragquas \\ xquad\_es} & 
\makecell[l]{arc\_ca \\ catalanqa \\ coqcat \\ openbookqa\_ca \\ piqa\_ca \\ siqa\_ca \\ xquad\_ca} & 
\makecell[l]{bertaqa \\ bhtc\_v2
 \\ eus\_exams \\ eus\_proficiency \\ eus\_trivia \\ vaxx\_stance
} & 
\makecell[l]{openbookqa\_gl} \\
\hline
\makecell[l]{Reading \\ comprehension} & 
\makecell[l]{belebele\_spa\_Latn} & 
\makecell[l]{belebele\_cat\_Latn} & 
\makecell[l]{belebele\_eus\_Latn \\ eus\_reading} & 
\makecell[l]{belebele\_glg\_Latn} \\
\hline
\makecell[l]{Summarization} & 
\makecell[l]{noticia \\ xlsum\_es} & 
\makecell[l]{cabreu} & 
\makecell[l]{--} & 
\makecell[l]{summarization\_gl} \\
\hline
\makecell[l]{Text \\ classification} & 
\makecell[l]{humorqa \\ fake\_news\_es \\ offendes} & % sad
\makecell[l]{catalonia\_ \\ independence} & 
\makecell[l]{bec2016\_eu} &
\makecell[l]{--}
\makecell[l]{} \\
\bottomrule
\end{tabular}
}
\caption{Datasets of \laleaderboard as of February 2025 organized by task type and language.} 
\label{tab:tasks_per_language}
\end{table*}

\begin{table*}[!ht]
\small
\centering
\scalebox{0.97}{
\begin{tabular}{p{2.5cm} p{3cm} p{8.2cm}}
\toprule
\makecell[l]{\textbf{Task Type}} & 
\makecell[l]{\textbf{Dataset}} & 
\makecell[l]{\textbf{Languages}} \\
\midrule
\makecell[l]{Adaptation} & 
\makecell[l]{\textit{phrases}} & 
\makecell[l]{Catalan, Spanish, Valencian} \\
\midrule
\makecell[l]{Common-sense \\ reasoning} & 
\makecell[l]{\textit{xstorycloze\_gl}} & 
\makecell[l]{Galician} \\
\midrule
\makecell[l]{Counter-narrative \\ generation} & 
\makecell[l]{\textit{conan\_eus/mt\_es} \\ \textit{refutes}} & 
\makecell[l]{Basque, Spanish \\ Spanish} \\
\midrule
\makecell[l]{Ethics} & 
\makecell[l]{\textit{h4rmony\_eval}} & 
\makecell[l]{Spanish} \\
\midrule
\makecell[l]{Text \\ classification} & 
\makecell[l]{\textit{haha}} & 
\makecell[l]{Spanish} \\
\midrule
\makecell[l]{Natural language} & 
\makecell[l]{\textit{americasnlp\_nli}} & 
\makecell[l]{Aymara, Asháninka, Bribri, Guarani, Náhuatl,\\ Otomí, Quechua, Rarámuri, Shipibo-Konibo, Wixarika} \\
\makecell[l]{inference} & 
\makecell[l]{\textit{meta4xnli}} & 
\makecell[l]{Spanish} \\
\midrule
\makecell[l]{} & 
\makecell[l]{\textit{paes\_cl}} & 
\makecell[l]{Spanish} \\
\makecell[l]{Question} & 
\makecell[l]{\textit{voces\_originarias}} & 
\makecell[l]{Aymara, Guarani, Tehuelche, Náhuatl, Quechua } \\
\makecell[l]{answering} & 
\makecell[l]{\textit{medexpqa}} & 
\makecell[l]{Spanish} \\
\makecell[l]{} & 
\makecell[l]{\textit{quales}} & 
\makecell[l]{Spanish} \\
\midrule
\makecell[l]{} & 
\makecell[l]{\textit{flores}} & 
\makecell[l]{Spanish, Catalan, Basque, Galician} \\
\makecell[l]{Translation} & 
\makecell[l]{\textit{americasnlp\_mt}} & 
\makecell[l]{Spanish, Aymara, Asháninka, Bribri, Guarani, Náhuatl,\\ Otomí, Quechua, Rarámuri, Shipibo-Konibo, Wixarika } \\
\makecell[l]{} & 
\makecell[l]{\textit{tradu\_latam}} & 
\makecell[l]{Spanish, Aymara, Guraraní, Tehuelche, Náhuatl, Quechua } \\
\midrule
\makecell[l]{Truthfulness} & 
\makecell[l]{\textit{truthfulqa}} & 
\makecell[l]{Spanish, Catalan, Basque, Galician} \\
\makecell[l]{} & 
\makecell[l]{\textit{veritasqa}} & 
\makecell[l]{Spanish, Catalan, Galician} \\
\bottomrule
\end{tabular}
}
\caption{Datasets that have been recently donated to \laleaderboard and are not yet included in the evaluation results, including benchmarks involving American Indigenous languages.}
\label{tab:upcoming_datasets}
\end{table*}

\paragraph{Languages} \laleaderboard contains 22 evaluation datasets in Spanish, including the varieties of Spain, Mexico, Argentina, Chile, and Uruguay. It also gathers datasets in all the official languages of Spain, with 18 datasets in Catalan, 17 in Basque, and 9 in Galician. 

\paragraph{Origin} We aim to evaluate models with high-quality datasets that reflect the cultural and linguistic idiosyncrasies of each language.
% LONG if needed, comment next sentence
For this reason, we only include datasets that have been annotated or revised by at least one native speaker of the language.
We prioritize the inclusion of datasets originally created in the language they evaluate, which constitute 55\% of the leaderboard. When this is not possible and translation is required, we prioritize datasets translated by human professionals. Not only does this prevent the loss of linguistic nuances that happens with machine translation \cite{lostintranslation}, but it also allows translators to adapt the text to the target culture \cite{nangia2020crows} and to identify errors in the source datasets and ensure that no extra hints regarding the answer are given in the input prompt \cite{baucells-etal-2025-iberobench}. In \laleaderboard, 38\% of the datasets have been manually translated from an existing English benchmark. We also acknowledge that, given the low-resource nature of some languages we cover, machine translation is more affordable than human translation. However, we only include such datasets if the automatic translation was comprehensively reviewed by a person proficient in the target language. Only 7\% of the datasets in \laleaderboard are manual reviews of machine-translated datasets.

\paragraph{Format} The multiple-choice question-answering (MCQA) format is widely used for automatic evaluations due to its simplicity. Thus, MCQA is the format of 59\% of the tasks included in \laleaderboard. We acknowledge that the literature has identified some issues with MCQA tasks, such as models' sensitivity to answer order \cite{pezeshkpour-hruschka-2024-large, mina-etal-2025-cognitive} or lack of task understanding \cite{khatun2024studylargelanguagemodels}. Moreover, some suggest that this type of task does not reflect the actual models' responses and capabilities \cite{li2024multiplechoicequestionsreallyuseful, wang-etal-2024-answer-c}. To address this issue, we also include text generation tasks, such as summarization, evaluated using NoticIA for Spanish, caBreu for Catalan, and Summarization-GL for Galician. We evaluate long-form question-answering
%(LFQA)
in Spanish using the AQuAS and RagQuAS datasets. Finally, we assess counter-narrative generation with RefutES in Spanish and CONAN in Basque and Spanish.

\paragraph{Domains} \laleaderboard includes well-known generalist datasets aimed at evaluating a model's capability to understand and complete a task, such as Belebele, WNLI, and XStoryCloze. We also include evaluation datasets focused on truthfulness assessment, such as VeritasQA and the Galician translation of TruthfulQA. There are, in addition, several domains represented in \laleaderboard, such as the medical (e.g., ClinTreatES), legal (e.g., SpaLawEx),
%political (e.g., VaxxStance),
and press (e.g., caBreu, NoticIA). We also include ethics-oriented datasets, evaluating stereotype generation in Spanish and Catalan with CrowsPairs. % and alignment with ecolinguistic values using H4rmonyEval.

\paragraph{Tasks}
The types of tasks chosen for our leaderboard extend those usually included in well-known leaderboards (e.g., reasoning, natural language inference, question answering or summarization) to other task types for which high-quality datasets exist in our target languages
(e.g., counter-narrative generation or linguistic acceptability).
For consistent performance comparisons across languages, we prioritize tasks available in multiple languages. % task parallelism
% The final set of task types included in \laleaderboard are:
% reasoning (commonsense and mathematical), natural language inference (NLI, including paraphrasing), question answering (QA), language proficiency (specifically reading comprehension, linguistic acceptability, and summarization), text classification, translation/adaptation, and alignment with ethical values (counter-narrative generation, stereotype generation, and ecolinguistics).

\subsubsection{Metrics}
\label{sec:metrics}
The MCQA tasks are evaluated by measuring the logarithmic probabilities ({\small \textsc{LogProbs}}) of models' outputs among a restricted list of options.
For text generation tasks, we compare the expected (\textit{gold-standard}) and given responses using various metrics depending on the original authors' implementation, including BLEU \citep{papineni-etal-2002-bleu}, ROUGE \citep{lin-2004-rouge} and Semantic Answer Similarity (SAS, \citealp{risch-etal-2021-semantic}). 
%To evaluate more complex and subjective tasks, we have also adapted an automated LLM-based metric from \citep{zubiaga2024llmbasedrankingmethodevaluation}.
Furthermore, following the recent trend of evaluating text generation tasks using LLMs, we are adapting an automated Judge-LLM metric from \citet{zubiaga2024llmbasedrankingmethodevaluation}. 
Since SAS and LLM-based metrics are not currently supported in the evaluation suite we use, the LM Evaluation Harness \citep{eval-harness}, we implement them in our open-source fork\footnote{\url{https://github.com/somosnlp/lm-evaluation-harness}}. %We hope to integrate these metrics into the official Harness repository soon.

\subsection{Code Bases}
\label{sec:code_bases}

\subsubsection{Backend}
\label{sec:backend}

We acknowledge the cost of running evaluations and want to ensure that any researcher or developer can compare their models to the state-of-the-art and follow their evolution. This is why submitting a model for evaluation is open to the whole community.
Once a model has been added to the evaluation queue, the last commit of the model is stored for reproducibility and to enable future comparisons of different versions.
The results from the LM Evaluation Harness \citep{eval-harness} are normalized according to the following formula:
\begin{equation}
  \resizebox{\columnwidth}{!}{%
    $\displaystyle
    \text{normalized\_value} = 
    \frac{\text{raw\_value} - \text{random\_baseline}}
         {\text{max\_value} - \text{random\_baseline}}
  $}
\end{equation}
where $random\_baseline$ is $0$ for generative tasks and $1/n$ for MCQA tasks with $n$ choices.

\subsubsection{Frontend}
\label{sec:frontend}

The implementation of \laleaderboard is based on the HuggingFace leaderboard template\footnote{\url{https://hf.co/spaces/demo-leaderboard-backend/leaderboard}}. The frontend is developed using Gradio \cite{abid2019gradio} and presents the evaluation results categorized by language.
To ensure transparency and reproducibility, we share the evaluation command and normalization formula.
%Additionally, we include a detailed acknowledgements section listing the contributors to the project.
To bring the tool closer to the
%target
community, the information and submission guidelines are available in English and Spanish.

\subsubsection{License}
\label{sec:license}

Since we want to motivate other communities to create their own, \laleaderboard is published under the permissive Apache 2.0 license\footnote{\url{https://www.apache.org/licenses/LICENSE-2.0}}.

\subsection{Efficiency Considerations}
\label{sec:efficiency_considerations}

\subsubsection{Number of Few-Shot Examples}
\label{sec:number_fewshots}
% advice to leaderboard builders? (Clementine)

Recent literature reveals significant inconsistency in the number of examples (shots) used when evaluating large language models (LLMs). While early research demonstrated notable performance improvements with 3-5 in-context examples \cite{brownetal2020}, current evaluation practices vary considerably across different models and benchmarks. For instance, the Open LLM Leaderboard employs 0-5 shots depending on the task, Mistral-7B generally follows this range with an exception of 8 shots for GSM8K \cite{cobbe2021trainingverifierssolvemath}, and Llama 3 and OLMo models focus primarily on zero-shot evaluation. In contrast, Gemini models use a broader range of 0-10 shots, including ``variable-shot'' configurations. This variation extends to language-specific models, with Salamandra\footnote{\url{https://hf.co/BSC-LT/salamandra-7b-instruct}} and Latxa \cite{latxa} families using different shot configurations in their evaluations, typically ranging from 0 to 5 shots.

Given this myriad of options, when choosing the number of shots to use in \laleaderboard, we take into consideration the following aspects:

\textbf{A. Base vs. instruct models} 
    %Base models benefit from few-shot evaluation \cite{brownetal2020}, but instructed models are expected to perform tasks in 0-shot settings, as their use is chat-oriented. Thus,
    The number of shots should allow for a fair evaluation of the base models without helping instruct models too much.
    Also, the availability of structured datasets in specific evaluation formats—such as MCQA—is very low in mid- and low-resource languages.
    %For instance, the number of MCQA datasets in Galician and Basque that can be used for training is extremely limited compared to the vast number of formatted datasets available in English.
    This means that models trained on English-heavy corpora are more likely to have encountered these structured formats in English than in other languages, potentially biasing their performance.
    
\textbf{B. Cognitive bias}
    Models suffer from cognitive bias depending on the order and options presented as few-shots \cite{zhao2021calibrateuseimprovingfewshot, pezeshkpour-hruschka-2024-large, mina-etal-2025-cognitive}. Thus, we ensure that, in MCQA tasks, all possible correct options are included in the in-context learning instances. For example, in an MCQA task with four possible answers, we evaluate on a 4-shot setting, with each shot showing one of the four options as correct, in random order. This is done unless it interferes with item A. %(e.g., if a task has too many options).
    
\textbf{C. Context windows} 
    The context window limitations of language models vary significantly based on hardware constraints and architectural choices, affecting their ability to process long-form tasks such as summarization and reading comprehension. For example, while the Spanish government's 40B-parameter ALIA model\footnote{\url{https://hf.co/BSC-LT/ALIA-40b}} operates with a 4,096-token context window, Meta's Llama 3.2 1B model can handle up to 128K tokens\footnote{\url{https://ai.meta.com/blog/llama-3-2-connect-2024-vision-edge-mobile-devices}}. To ensure fair evaluation across models with different context window capacities, few-shot examples are employed with a maximum limit of 2,048 tokens, following the methodology established in previous research on LLM analysis \cite{biderman2023pythiasuiteanalyzinglarge}.
    
\textbf{D. Prompt format}
    %During evaluation, we use the existing prompt available for each task in the LM Evaluation Harness. For tasks not already implemented in the Harness, we create prompts following a similar format to other tasks of the same type and verify them with the original authors of each dataset. If a prompt is overly complex or convoluted (e.g., for paraphrasing tasks such as PAWS and reasoning such as COPA), we provide 3 in-context examples to the model, following \citet{brownetal2020}'s suggestion that it should allow the models to understand the task while complying with the aspects discussed in items A and C. Similarly, when a prompt is straightforward and includes answers (as in most QA tasks), we use a 2-shot evaluation, again allowing base models to understand the simple format. In contrast, for the tasks with a very explicit prompt that uses a naturally occurring structure and language (e.g., those in ClinDiagnosES and NoticIA), we carry out 0-shot evaluations. The same applies to the tasks evaluated by looking at the probability of a sentence being generated as a continuation of the prompt (e.g., XStoryCloze).
    The evaluation methodology employed task-specific prompts from the LM Evaluation Harness, with new prompts created for previously unimplemented tasks following established formats and validated by dataset authors. The number of few-shots varied based on prompt complexity: convoluted prompts (e.g., paraphrasing with PAWS and reasoning with COPA) used 3 in-context examples to allow models to understand the task while complying with items A and C \cite{brownetal2020}; straightforward question-answering tasks employed 2-shot evaluation, while tasks with explicit, naturally structured prompts (like ClinDiagnosES and NoticIA) and those evaluating sentence continuation probability (e.g., XStoryCloze) were conducted using 0-shot evaluation.

%%% TABLA 3

\subsubsection{Measuring Model Efficiency}
\label{sec:measuring_model_efficiency}

% Detalles de la máquina los tienes aquí: https://www.res.es/es/nodos-de-la-res/marenostrum-5 (mira la parte ACC)
% De momento las H100 no están incluidas en https://mlco2.github.io/impact/ 

The evaluation was performed using two NVIDIA H100 GPUs with Hopper architecture and 64 GB of HBM memory in the MareNostrum 5 High-Performance Computer\footnote{\url{https://www.bsc.es/ca/marenostrum/marenostrum-5}}, maintaining identical configurations across instances to ensure consistent measurements. Performance metrics included task execution time and energy consumption, tracked using the Energy Aware Runtime (EAR) package\footnote{\url{https://www.bsc.es/research-and-development/software-and-apps/software-list/ear-energy-management-framework-hpc}}, with all tasks running at a batch size of 1.

Task execution duration, which includes token prediction time, response length, and tokenizer efficiency, was measured to assess model speed for time-sensitive applications. The duration of task execution is influenced by multiple factors beyond token prediction time, including the response length generated and the language-specific tokenization efficiency \cite{10632720}.

Energy consumption was recorded in kWh and converted to CO$_2$ equivalents using the European Commission's conversion ratio for Spain (0.158 kg CO$_2$/kWh), as the evaluation was conducted in Barcelona \cite{lottick2019energy}. %The total computational resources amounted to 660.87 hours of processing time and 582.84 kWh of energy consumption, resulting in 92.09 kg of CO$_2$ emissions.

\begin{table}[!t]
    \centering
    \scriptsize
    \scalebox{0.93}{
\begin{tabular}{lccccc}
    \toprule
    \textbf{Model} & \textbf{Top10} & \textbf{ES} & \textbf{CA} & \textbf{EU} & \textbf{GL} \\
    \midrule
    gemma-2-9b-it & \textbf{43} & 61.69 & 57.30 & \textbf{54.13} & 46.49 \\
    gemma-2-9b & \textbf{43} & 57.21 & \textbf{59.60} & 53.80 & 48.58 \\
    Meta-Llama-3.1-8B-IT & 41 & 59.03 & 57.01 & 49.87 & 45.07 \\
    Qwen2.5-32B-IT-GPTQ-Int4  & 38 & \textbf{64.06} & 56.80 & 49.23 & \textbf{52.52} \\
    Qwen2.5-14B-IT-GPTQ-Int8 & 35 & 60.59 & 54.08 & 49.05 & 52.13 \\
    EuroLLM-9B \textbf{*} & 30 & 55.00 & 57.32 & 38.92 & 46.36 \\
    Meta-Llama-3.1-8B & 28 & 55.62 & 56.52 & 46.90 & 44.90 \\
    salamandra-7b \textbf{*} & 26 & 52.17 & 54.13 & 45.80 & 39.88 \\
    salamandra-7b-instruct \textbf{*} & 25 & 51.41 & 53.22 & 46.19 & 41.65 \\
    Qwen2.5-7B  & 25 & 58.79 & 57.28 & 42.51 & 46.82 \\
    aya-expanse-8b & 20 & 55.42 & 53.99 & 41.99 & 43.38 \\
    %rigochat
    %mistral
    %occiglot
    %EuroLLM-9B-IT \textbf{*} & 18 & 57.21 & 52.96 & 38.00 & 44.47 \\
    %Qwen2.5-7B-IT & 18 & 57.46 & 48.20 & 41.36 & 43.13 \\
    %Yi-1.5-9B & 13 & 54.51 & 54.17 & 40.36 & 44.44 \\
    %occiglot-7b-eu5 \textbf{*} & 13 & 55.02 & 53.71 & 38.73 & 45.62 \\
    \midrule
    salamandra-2b \textbf{*} & \textbf{8} & 46.60 & 41.05 & 39.28 & 34.99 \\
    Qwen2.5-1.5B & \textbf{8} & 52.60 & 47.88 & 38.13 & 38.79 \\
    gemma-2-2b-it & 5 & \textbf{54.85} & 48.27 & \textbf{50.27} & \textbf{39.79} \\
    Qwen2.5-1.5B-IT & 5 & 54.38 & 45.09 & 39.04 & 39.27 \\
    salamandra-2b-instruct \textbf{*} & 3 & 43.83 & 43.28 & 38.57 & 34.53 \\
    gemma-2-2b & 2 & 51.59 & \textbf{50.22} & 39.81 & 39.29 \\
    leniachat-qwen2-1.5B-v0 \textbf{*} & 2 & 51.14 & 40.97 & 38.29 & 36.16 \\
    EuroLLM-1.7B-IT \textbf{*} & 2 & 47.92 & 45.28 & 36.19 & 34.67 \\
    %FLOR-1.3B-IT \textbf{*} & 40.31 & 46.23 & 44.68 & 37.14 & 33.17 \\
    %SmolLM2-1.7B & 1 & 47.61 & 41.58 & 36.80 & 34.73 \\
    \bottomrule
\end{tabular}
}
    \caption{
    Best performing models ranked by the number of tasks in which they achieved a top-10 placement (see “Top10” column), showing their average language performance. Target language–optimized models are indicated by an asterisk (*). Complete per-language top-10 frequencies can be found in Figure~\ref{fig:barplots_top10_frequency_by_language}, and full average scores for all models are shown in Figure~\ref{fig:heatmap_models_avg_scores}.
    }
    \label{tab:model_comparison}
\end{table}

%%% FIGURE 2

\section{Evaluation Results and Analysis}
\label{sec:evaluation_results}

We focus on evaluating models accessible to the broader community. We select 50 open-weights models from various families, primarily ranging from 1B to 9B parameters, while including larger quantized models.  We consider multilingual models trained by large technological companies and startups like
% Google-Gemma \cite{team2024gemma2},
% Alibaba-Qwen \cite{qwen25}, 
Meta-Llama
%(from Meta)
\cite{llama3} and Mistral \cite{mistral7b}
as well as models such as EuroLLM \cite{martins2024eurollmmultilinguallanguagemodels} and Salamandra\footnote{\url{https://hf.co/collections/BSC-LT/salamandra-66fc171485944df79469043a}}, which
% FLOR \cite{da-dalt-etal-2024-flor}, Gromenauer\footnote{\url{https://hf.co/bertin-project}}, Occiglot\footnote{\url{https://hf.co/collections/occiglot/occiglot-eu5-7b-v01-65dbed502a6348b052695e01}} or EuroLLM among others. These models
have been designed by European and Spanish research groups specifically to process our target languages more efficiently and capture cultural nuances. We assess both the base and instruction-tuned versions when available (Appendix~\ref{sec:appendix_models_evaluated}).
Each model-task pair was evaluated once, and the raw, pre-normalization results analyzed in this work are publicly available\footnote{\url{https://hf.co/datasets/la-leaderboard/results}}.

\subsection{Model performance}

% Evaluation results: https://hf.co/datasets/la-leaderboard/results-bsc/tree/main/0-results/results (one JSON per model).  

Table~\ref{tab:model_comparison} shows that the models most consistently ranking in the top 10 across languages are Gemma-2-9B base and instructed, Llama-3.1-8B-IT, and the quantized versions of Qwen-2.5-IT 14B and 32B. When prioritizing transparency, EuroLLM-9B stands out, followed by Salamandra-7B. For resource-constrained settings, Salamandra-2B and Qwen-2.5-1.5B offer the best performance. % among smaller models.

\begin{figure}[!t]
    \centering
  \includegraphics[width=0.9\columnwidth]{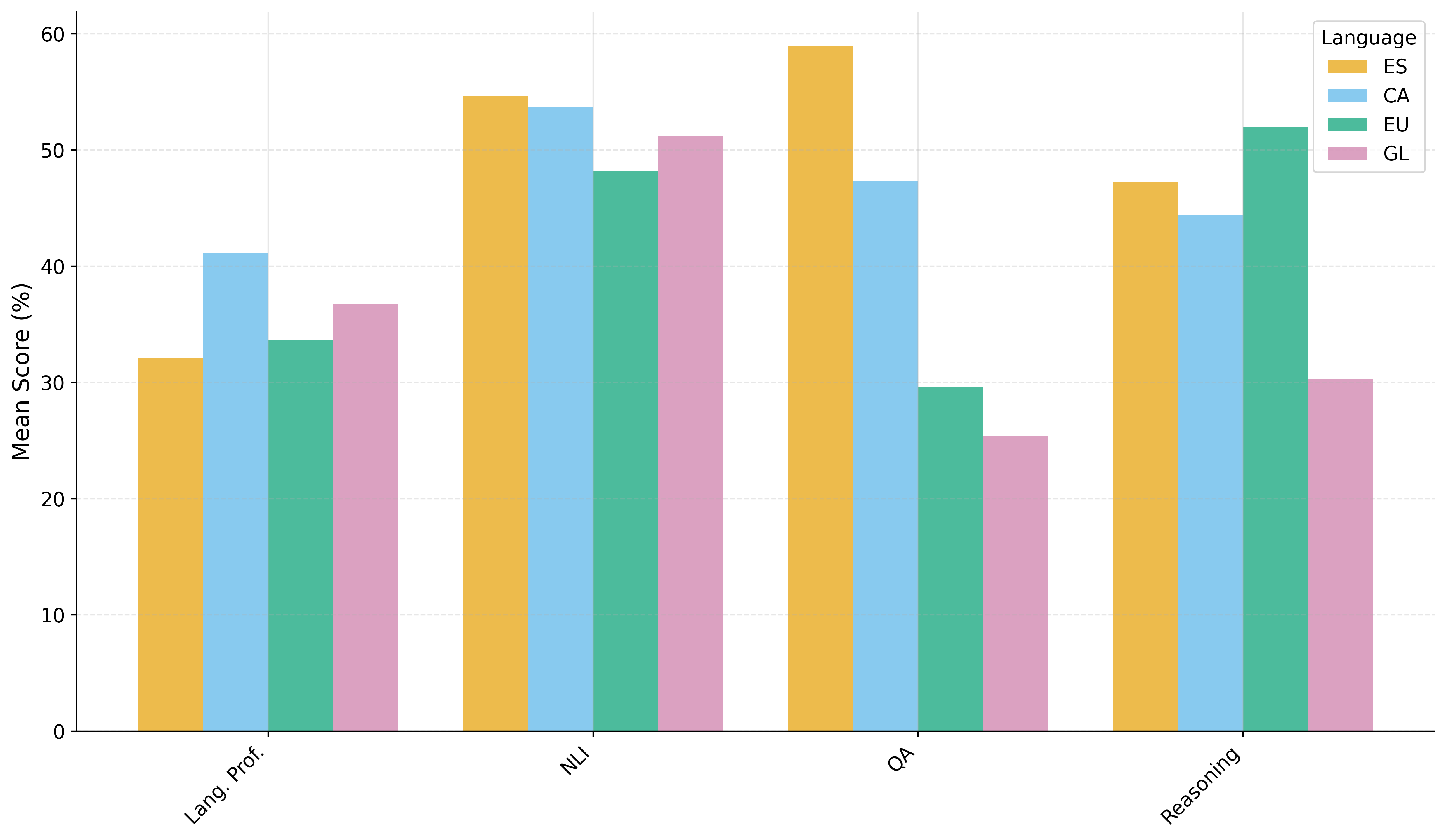}
  \caption{Results per type of task type and language.} %, where "Language Proficiency" includes reading comprehension, linguistic acceptability and summarization, "NLI" includes textual entailment and paraphrasing, and "Reasoning" includes commonsense and mathematical reasoning.}
\label{fig:barplot_performance_per_task_type_main}
\end{figure}

\paragraph{Performance per task}
% FUTURE WORK: This would require a per-sample analysis to try to understand the trends (Clem)
As illustrated in Figure~\ref{fig:barplot_performance_per_task_type_main}, the evaluation results are generally better for NLI tasks, including textual entailment and paraphrasing, and worse for language proficiency tests, with all four languages having similar performance on both tasks. Within the language proficiency tests, which combine reading comprehension, linguistic acceptability, and summarization, results are particularly low for summarization tasks.
% FUTURE WORK: Do language-optimized models perform better at these reading comprehension tasks? (Antoine)
In question answering and reasoning tasks, there is a larger inter-language difference, with Galician having significantly lower scores overall, while Basque obtains the best results for reasoning but the second worst for question answering. 
While commonsense reasoning results are generally good, math reasoning yields the lowest results, which could be related to a too strict metric (exact match).
Further analysis is needed to understand whether these differences are due to the datasets used or indeed to the models' performance. The poor results on language proficiency tests also warrant a detailed examination to understand their implications, as they may indicate fundamental limitations in the models' knowledge across languages.

\paragraph{Performance per language} 
% Párrafo eliminado porque creo que comparar resultados entre idiomas no es correcto, las tareas no son paralelas
% In general, results are better for Spanish and Catalan and worse for Basque and Galician. This was expected for Basque, a language isolate (therefore very different from the other languages of the leaderboard), but not fully for Galician, as it shares Latin roots with Spanish and Catalan. However, the generalized lower scores in Galician could be a consequence of the reduced number of training and instruction datasets available for this language.

Regarding specific models, the same five (Gemma2, Llama3.1, and quantized Qwen2.5) consistently rank among the top ten performers. However, we find that some models stand out for specific language-task pairs.
In Spanish, AYA-Expanse-8B joins the top three for QA, EuroLLM-9B for reasoning, and RigoChat-7B is in the top four for language proficiency.
In Catalan, EuroLLM ranks among the top three for both QA and reasoning tasks.
For Basque, Salamandra-7B-IT is the second-best QA model, while Latxa-7B and Salamandra-7B occupy the first and second positions in reasoning and the third and second positions in classification, respectively.
In Galician, EuroLLM-9B is the best model for QA, followed by its instructed version, Salamandra-7B ranks third in reasoning, and AYA-Expanse-8B takes second place in NLI.% (Figure~\ref{fig:barplots_top10_frequency_by_language}).

%\paragraph{SOTA vs. Language-optimized models}
%We observe that the first two-thirds of the top 15 models are SOTA models. This distribution suggests that technological advances in state-of-the-art models, coupled with access to greater resources by the companies involved in training them, play a more decisive role in the performance of language models than any specific solution based on pre-training, fine-tuning, or other mechanisms.
% Discuss in relation to compute budget, training time and training data size

\paragraph{Performance by training data configuration}

It is valuable to compare the different language configuration strategies in training.
Salamandra and EuroLLM, by 
distributing tokens in the pre-training data fairly among languages, achieve homogeneous performance and, in particular, reinforce Galician and Basque, but lose ground in Spanish and Catalan to models supported by high-resource corpora.
Gemma2, which prioritizes volume over linguistic diversity, stands out for just the opposite; training on a large but linguistically imbalanced dataset allows the model to transfer patterns learned from English, maths, and code, and consistently rank among the top models overall.
This shows that, sometimes, breadth of knowledge can compensate for the lack of language-specific data.

The results confirm that the strategy of pre-training from scratch on a large multilingual corpus is the one that offers the most consistent and homogeneous coverage. Qwen-2.5, trained on 29 languages, and Llama-3.1, trained on 8, rank high in all languages.
Moreover, systems that use continuous pre-training - adding large-scale data in a new language plus some English data to avoid catastrophic forgetting - achieve peak scores in the target language; e.g., Latxa holds the second-highest mark in Basque reasoning.
Finally, models that just fine-tune on target language data without extending the pre-training (e.g., RigoChat or some Instruct variants) gain fluency and style, but the improvement in reasoning and QA tasks is modest, and models rarely enter the top 10.
In summary, the earlier and deeper the model is exposed to the target languages, the higher its average scores; a posteriori strategies function as beneficial reinforcers, especially in less-resourced languages, but do not replace the power of extensive multilingual pre-training.

% Gemma2: Web documents ("Primarily English-language content"), code and mathematical text
% Qwen 29 lang (inc. French, Spanish, Portuguese, Italian)
% Llama 8 lang (English, German, French, Italian, Portuguese, Hindi, Spanish, and Thai)
% Salamandra pre-trained 35 European languages and 92 programming languages
% EuroLLM pre-trained

% FUTURE WORK
% Con respecto a la sección training data configuration, sería interesante comentar que algunos de los "language-specific models" han sido pre-entrenados from scratch (salamandra, por ejemplo), mientras que otros (latxa, por ejemplo) han hecho continual pre-training intentando no olvidar el inglés (catastrophic forgetting). Esto + la cantidad de texto usado (no sé si este dato está disponible) por cada idioma y cómo se introduce conocimiento lingüístico durante el pre-entrenamiento es clave para el buen rendimiento de modelos como gemma, claro (ya sé que igual estoy "stating the bleeding obvious", pero esto todavía no está muy claro cómo hacer para obtener buenos modelos multilingües, sobre todo para idiomas con menos recursos). También la generación de instrucciones y el post-training en los idiomas de destino es interesante comentar, pero no sé si da para tanto.

\paragraph{Performance by training compute}

Regarding the computational budget and the hardware used, public data is scarce. 
Regarding large tech companies we can consider Llama-3.1-8B, which was trained for 1.5M GPU-h. By contrast, other evaluated models required three to four orders of magnitude fewer. As expected, the results confirm that teams with access to GPUs/TPUs and high-end accelerators (TPUv5p, H100 or A100 in large clusters) can apply the strategies that we highlighted in the previous section as the most successful - massive multilingual pre-training and continuous pre-training - reflecting in significantly higher average scores. However, the same results also nuance the linear relationship between investment and performance
%: EuroLLM-9B achieves strong results with just 4 k h-GPU thanks to thorough corpus curation, while
as Salamandra-7B, despite its large budget (600 k GPU-h), falls in the middle of the table. Overall, the dashboard illustrates that computational abundance enables impactful techniques, but that dataset quality and fine-tuning remain decisive levers for converting that power into real performance.

\paragraph{Performance vs. size}
In general, our experiments show some correlation between performance and size, with models in the range of 1-2B parameters achieving better scores for their size. This is particularly true for Gemma-2-2B and Qwen-2.5-1.5B, both base and instructed models.
Among the top 10 models, we find that all have between 8 and 9 billion parameters, except for the quantized versions from the Qwen family.
% Figure~\ref{fig:scatter_plot_performance_vs_size_all_languages}

\paragraph{Performance per model type}
For the same VRAM requirement, quantized models with more parameters outperform full-precision models with fewer parameters.
We don't observe any correlation between performance and whether the model is pre-trained or instruction-tuned, as this varies depending on the model. 
%% Models not optimized for the target languages (e.g., Gemma) achieve the highest scores, while fine-tuned or continually pre-trained models on these languages (e.g., EuroLLM) outperform foundation models designed with the same linguistic focus (e.g., Salamandra).

\subsection{Energy consumption}
The total computational resources amount to 660.87 hours of processing time and 582.84 kWh of energy consumption, resulting in 92.09 kg of CO$_2$ emissions. 
On average, each model consumes 9.25 kWh (median 6.88, SD 8.42), showing a wide variety in energy usage.
The models that consume the most energy are Grommeanuer-7B-IT, Qwen-2.5-32B-IT-GPTQ-Int4, and Qwen-2.5-14B-IT-GPTQ-Int8, each exceeding 30 kWh. On the other hand, Salamandra-2b, FLOR-1.3B-IT, and LLama-3.2-1B-IT are the most energy-efficient, consuming less than 2.1 kWh each. 
% Figure~\ref{fig:carbon_emissions_total} represents the total energy consumed by each model.

\paragraph{Energy consumption vs. size}
%As expected, the general trend indicates that larger models consume more energy, with consumption increasing approximately threefold between the smallest models (1–2 billion parameters) and the largest ones (6–9 billion).
As expected, a strong correlation between model size and energy dissipation is observed, as the number of arithmetic operations required to predict a token is related to the number of parameters of the model. 
However, some outliers are observed, such as Qwen, which consumes significantly more energy across all its sizes compared to other models. Conversely, models like FLOR exhibit considerably lower energy consumption across their different sizes relative to other models of similar scale. % REVIEW
% Figure~\ref{fig:carbon_emissions_vs_size} presents a comparison between model size and energy consumption. 

\paragraph{Energy consumption vs. task}
Similarly, as anticipated, text generation tasks such as summarization require more energy for evaluation.
Since LLMs generate text token by token and their prediction speed remains constant (assuming the same hardware and stable conditions), the most energy-intensive tasks are expected to be those that require the generation of larger amounts of text.

\paragraph{Energy consumption vs. performance}
Our experiments show a strong correlation between the energy consumed at inference and the model performance. For one of the overall top models, Gemma-2-9B, its instruction-tuned version excels with a third of the energy consumed by the base version. We observe a trend of instructed models consuming less energy, due to the verbosity of base models. 
% Figure~\ref{fig:performance_vs_energy_kwh}

%Figures \ref{fig:carbon_emissions_total}, \ref{fig:carbon_emissions_by_task} and \ref{fig:carbon_emissions_vs_size} in Appendix~\ref{sec:appendix_evaluation_results} provide more details about the efficiency of the models evaluated.

\section{Conclusions and Future Work}
\label{sec:conclusion}

In this paper, we propose a methodology to create community-driven leaderboards, including
%an open data collection campaign,
key points to gather diverse datasets and the rationale behind a more efficient and accessible evaluation setup. In doing so, we hope to inspire the creation of more leaderboards that fulfil the needs of diverse linguistic communities.

In particular, we present \laleaderboard, the first open-source leaderboard to evaluate LLMs in languages from Spain and Latin America. It is the result of a collaboration among 13 research groups.
%from various regions in Spain and Latin America.
\laleaderboard consists of 66 datasets in Spanish, Catalan, Basque, and Galician, and covers a wide range of task types and domains.
The results of evaluating 50 LLMs suggest that the top-performing models in the target languages are Gemma-2-9B base and instruct, Llama-3.1-8B-IT, and the quantized versions of Qwen-2.5-IT 14B and 32B. However, for domain-specific applications, particular linguistic contexts, or deployment scenarios with computational constraints or transparency requirements, alternative models (e.g., Salamandra-7B and EuroLLM-9B) demonstrate competitive performance metrics. 
%Nonetheless, when considering aggregate performance across the complete spectrum of evaluated languages and tasks, models subjected to extensive multilingual pre-training 
%While for specific tasks and languages, or when taking into account low-resource computing requirements or criteria such as transparency, some other models (e.g., Salamandra or EuroLLM), perform competitively, in terms of overall performance across all languages and tasks, models pre-trained with extensive multilingual data would remain the preferred choice. % continuar enfocándonos en "recomendaciones" de qué modelos utilizar
% O algo como "After evaluating 50 models, we provide recommendations to choose models depending on the language-task pair. For example, ...
% TODO

Our planned next steps include evaluating the recently donated datasets, with a special focus on indigenous languages. We will also add larger open models and proprietary models.
Moreover, we are organizing a hackathon to create a benchmark to measure cultural adequacy in each Spanish-speaking country.
%Multimodal to leverage another community-based dataset CVQA.
Finally, we welcome any person or organization interested in joining our effort.
This way, we hope that \laleaderboard will keep evolving to include more languages, language varieties, and use cases that motivate the development of LLMs that better serve our diverse community.

%%% 8 pages until HERE %%%
% If accepted, we have 1 extra page
% Unlimited space for "Limitations" and "Ethical considerations"

\section*{Limitations}

\paragraph{Indigenous languages}
We acknowledge that indigenous languages from Latin America are not yet included among the evaluation results of \laleaderboard.
% To the best of our knowledge, the only datasets that could be adapted to evaluate generative LLMs in indigenous languages are the ones from the shared tasks of AmericasNLP \cite{mager2021findings, ebrahimi2021americasnli}...
However, we have ongoing collaborations to include existing benchmarks and create new ones to keep extending \laleaderboard to be as inclusive as possible and reflect the diversity of the Spanish-speaking community.

\paragraph{Spanish language varieties}
Currently, the leaderboard includes datasets in the Spanish varieties of Spain, Mexico, Argentina, Chile, and Uruguay. Although we don't know the exact origin of all the samples from some third-party datasets, we estimate that less than 25\% of all the Spanish datasets in the leaderboard come from LATAM. We plan to increase this percentage by collaborating with LATAM research groups in the creation of an open hackathon.

\paragraph{Large and proprietary models}
To improve the coverage of the state-of-the-art language models for the use cases included in \laleaderboard, it would be interesting to evaluate larger language models
%, in the range of 70 billion parameters,
as well as proprietary models.

\paragraph{Contamination}
Another pending task is to analyse potential contamination \cite{sainz2023nlp} within our leaderboard. We have not addressed this yet because a high percentage of the datasets used are very recent and niche, making it unlikely that they have been incorporated into training data, unlike more established benchmarks such as MMLU \cite{MMLU, mmlu-pro, mmlu-pro+} that serve as primary pillars in model evaluation in every model report. Nevertheless, we have started to evaluate contamination to ensure in the short-term future that we provide high-quality results.

For the datasets specifically created for \laleaderboard, we advised the corresponding authors to release them gated to avoid being unintentionally included in training datasets by web scraping; AQuAS and RagQuAS are gated. The authors of TELEIA decided to release an adaptation of their dataset and keep the original private to be able to analyze contamination through time.

\section*{Ethical Considerations}

\paragraph{Fair representation}
Since our objective is to establish an evaluation standard for Latin America and Spain, it is important to properly represent the linguistic and cultural diversity of the community in order to avoid the perpetuation, or even amplification, of stereotypes and inequalities.
% mager2023ethical (https://arxiv.org/pdf/2305.19474)

\paragraph{Third-party datasets}
Some of the evaluation datasets included in \laleaderboard were created by organizations other than our data contributors.
As a result, we acknowledge the possibility that some of these datasets may have been developed using practices that could be considered unethical. These concerns range from potential legal violations to extractive data collection methods that may impact disadvantaged communities.

\paragraph{Environmental impact}
Evaluating 50 language models on 66 tasks required 660.87 hours of compute, translating to 92.09 kg of CO$_2$. However, we hope that by publishing a comprehensive evaluation of the available models, \laleaderboard will contribute to reducing the total environmental impact of individual private evaluations.

\paragraph{Misuse of La Leaderboard}
We welcome model submissions from everyone. This could potentially lead to overuse, with people sending many different versions of the same model. We plan to mitigate this behaviour by following the spam mitigation strategies from the Open LLM Leaderboard \cite{open-llm-leaderboard-v2}. % and requiring manual approval for models larger than XXB. 

\section*{Acknowledgments}

\laleaderboard is the result of the collaboration between many researchers, and we extend our heartfelt thanks to each of them.
% Example from BigScience paper
%Contributors are assigned to each category according to which aspects of the project they contributed to.
%Many authors appear under multiple categories because they contributed to the project in more than one way.
%Author order in all categories is alphabetical by first name.
%, except for “Major Contributors” where authors are shuffled randomly apart from Teven Le Scao, who is intentionally listed first and “Organization” where Thomas Wolf is intentionally listed last. A description of each category follows. For details on finer-grained contributions, please see the dataset papers.

\paragraph{Coordination}
María Grandury,
Javier Aula-Blasco,
Clémentine Fourrier,
Marta Guerrero,
Pedro Reviriego.

\paragraph{Dataset donation leads} Javier Aula-Blasco, Omar Sanseviero, Alejandro Vaca, Marta Guerrero, Álvaro Barbero, Rodrigo Agerri, Javier Conde, Flor Miriam Plaza-del-Arco, María Teresa Martín-Valdivia, Helena Gómez, Luciana Benotti, Iker García-Ferrero, Jorge Vallego, Luis Chiruzzo.

\paragraph{New dataset creation and annotation} 
Carmen Muñoz, Helena Montoro, Leire Rosado, Marta Guerrero, Nuria Aldama, Natàlia Fuertes, % IIC
Alejandro Vaca, % LenguajeNaturalAI
Marina Mayor-Rocher, Nina Melero, Elena Merino-Gómez, Miguel González, Raquel Ferrando, Javier Conde, Gonzalo Martínez, Pedro Reviriego % UPM
Daniel Vila, Álvaro Bartolomé,  Ignacio Talavera. % Argilla

\paragraph{Metric and task implementation}
Gonzalo Santamaría, David Betancur, % metrics
Anna Sallés, Irene Baucells, Javier Aula-Blasco, Julen Etxaniz, Alejandro Vaca, María Grandury, Gonzalo Martínez, Miguel González, Iker García-Ferrero, Guido Ivetta. % datasets
María Estrella Vallecillo-Rodríguez, Irune Zubiaga. % contranarrativa

\paragraph{Leaderboard implementation}
María Grandury, % SomosNLP
Clémentine Fourrier, Nathan Habib. % HF

\paragraph{Model evaluation}
Júlia Falcão,
María Grandury, Miguel González, Gonzalo Martínez, Javier Aula-Blasco, Bram Vanroy.

\paragraph{Paper writing} 
All the co-authors, especially
María Grandury,
Rodrigo Agerri,
Pedro Reviriego,
Javier Aula-Blasco,
Flor Miriam Plaza-del-Arco,
Clémetine Fourrier,
Gonzalo Martínez,
Luis Chiruzzo,
Alejandro Vaca,
Carmen Muñoz,
Marta Guerrero,
Javier Conde,
Guido Ivetta,
Júlia Falcão,
Álvaro Barbero,
Manuel Mager.

\paragraph{Communication}
Maria Sayavera, Jaume Lozano, % BSC
Florent Daudens, Brigitte Tousignant, % HF
Rafael Muñoz, Valentín Cardeñoso. % SEPLN

\paragraph{Computational resources}
Barcelona Supercomputing Center, Universidad Politécnica de Madrid, HuggingFace, The Flemish Supercomputer Center.

\paragraph{Funding} The work of the co-authors of this paper has been supported by multiple projects.
\begin{itemize}
    \item The coordination and implementation of \laleaderboard by María Grandury was carried out on a voluntary basis for six months, after which her work was supported by the Universidad Politécnica de Madrid (UPM).
    \item The work by the UPM team was supported by the Agencia Estatal de Investigación (AEI) (doi:10.13039/501100011033) under Grants FUN4DATE (PID2022-136684OB-C22) and SMARTY (PCI2024-153434) and by the European Commission through the Chips Act Joint Undertaking project SMARTY (Grant 101140087).
    \item The work by the Barcelona Supercomputing Center team and the datasets donated by ILENIA are funded by the Ministerio para la Transformación Digital y de la Función Pública and Plan de Recuperación, Transformación y Resiliencia - Funded by EU – NextGenerationEU within the framework of the project ILENIA with references 2022/TL22/00215337, 2022/TL22/00215336, 2022/TL22/00215335 and 2022/TL22/00215334, and within the framework of the project Desarrollo Modelos ALIA.
    \item The work of LenguajeNatural.AI is self-funded and self-supported by the company’s resources. %, with no contribution from public administrations or private funding.
    \item The contributions of Rodrigo Agerri and Irune Zubiaga are partially funded by the following MCIN/AEI/10.13039/501100011033 projects: (i) DeepKnowledge (PID2021-127777OB-C21) and ERDF A way of making Europe; (ii)  DeepMinor (CNS2023-144375) and European Union NextGenerationEU/PRTR. Irune Zubiaga is supported by a UPV/EHU PIF 2025 grant.
    \item The contributions of Guido Ivetta are supported by Fundación Vía Libre and Universidad Nacional de Córdoba, Argentina.
    \item The work by the SINAI team has been partially supported by projects CONSENSO (PID2021-122263OB-C21), MODERATES  (TED2021-130145B-I00), SocialTOX (PDC2022-133146-C21) funded by Plan Nacional I+D+i from the Spanish Government.
    \item The contributions of Jorge Vallego are supported by the H4rmony project and the Asociación Latinoamericana de Ecolingüística.
\end{itemize}

\paragraph{Community}
This project is a community-driven effort that continues to evolve. We thank everyone who has provided valuable feedback via ACL Rolling Review (ARR), the SomosNLP Discord community, the interface's Discussions tab, and in person at the KHIPU 2025 conference and the NAACL 2025 workshop "Language Models for Underserved Communities". We are grateful for the support, and invite the entire Ibero-American NLP community to join us.

\clearpage

\bibliography{main}

\appendix

\clearpage
\section{Evaluation Datasets}
\label{sec:appendix_datasets}

The datasets are used only for evaluation, aligning with their intended uses.

\subsection*{Spanish datasets}

The Spanish datasets in \laleaderboard are:
AQuAS \citep{iic2024aquas},
Belebele \citep{bandarkar-etal-2024-belebele},
EsCoLA \citep{bel-etal-2024-escola-spanish},
Fake News ES \cite{fake_news_es},
FLORES-200 \citep{nllbteam2022language},
ClinTreatES and ClinDiagnosES \citep{lenguajenaturalai2024medicalexpertes},
HumorQA \citep{lenguajenaturalai2024humorqa},
MGSM \citep{shi2023language}, 
MultiLingualCrowsPairs \cite{nangia2020crows},
% https://gitlab.inria.fr/corpus4ethics/multilingualcrowspairs
NoticIA \citep{garcíaferrero2024noticiaclickbaitarticlesummarization},
OffendES \cite{plaza-del-arco-etal-2021-offendes},
RagQuAS \citep{iic2024ragquas},
SpaLawEx \citep{lenguajenaturalai2024spalawex},
TELEIA \citep{spanish_benchmark_teleia},
WNLI \cite{gonzalez-agirre-etal-2024-building-data, baucells-etal-2025-iberobench}\footnote{For Spanish, see \url{https://hf.co/datasets/PlanTL-GOB-ES/wnli-es}.},
XL-Sum \citep{hasan-etal-2021-xl},
XStoryCloze \citep{lin-etal-2022-shot, baucells-etal-2025-iberobench},
and
XQuAD \citep{artetxe-etal-2020-cross}.

\subsection*{Catalan datasets}

The Catalan datasets in \laleaderboard are:
caBREU, CatalanQA, COPA-ca, CoQCat, PAWS-ca, TE-ca, WNLI-ca and XNLI-ca \citep{gonzalez-agirre-etal-2024-building-data},
IberoBench \citep{baucells-etal-2025-iberobench}, 
CatCoLA \citep{catcola2024}, 
FLORES-200 \citep{nllbteam2022language},
MGSM \citep{shi2023language}, 
XStoryCloze \citep{lin-etal-2022-shot},
XQuAD-ca \citep{armengol-estape-etal-2021-multilingual}, 
XStoryCloze \citep{lin-etal-2022-shot, baucells-etal-2025-iberobench},
Parafraseja\footnote{\url{https://hf.co/datasets/projecte-aina/Parafraseja}}, PAWS-X \citep{yang-etal-2019-paws}, 
and
VeritasQA \citep{aula-blasco-etal-2025-veritasqa}.

\subsection*{Basque datasets}

The Basque datasets in \laleaderboard are:
EusExams, EusReading, EusProficiency and EusTrivia from \citet{latxa};
BEC2016eu, BHTCv2, EpecKorrefBin, QNLIeu, WiCeu from BasqueGlue \cite{urbizu-etal-2022-basqueglue};
QNLI-eu \citep{urbizu-etal-2022-basqueglue}, 
VaxxStance \cite{VaxxStanceIberLEF2O},
XNLIeu \citep{heredia-etal-2024-xnlieu},
FLORES-200 \citep{nllbteam2022language},
MGSM \citep{shi2023language}, 
and
XStoryCloze \citep{lin-etal-2022-shot, baucells-etal-2025-iberobench}.

\subsection*{Galician datasets}

The Galician datasets in \laleaderboard are:
FLORES-200 \citep{nllbteam2022language},
GalCoLA \citep{de-dios-flores-etal-2023-dependency}, 
TruthfulQA-GL\footnote{\url{https://hf.co/datasets/proxectonos/truthfulqa_gl}},
and
XStoryCloze \citep{lin-etal-2022-shot, baucells-etal-2025-iberobench}\footnote{For Galician, see \url{https://hf.co/datasets/proxectonos/xstorycloze_gl}.}.

\subsection*{Datasets created for La Leaderboard}

The 7 datasets specifically created for the initial version of \laleaderboard are AQuAS, ClinDiagES, ClinTreatES, HumorQA, RagQuAS, SpaLawEx, and TELEIA. Their corresponding datasheets are included in Appendix~\ref{sec:appendix_datasheets}.

\subsection*{Newly donated datasets}

The new datasets donated will be evaluated shortly. These include CONAN-EUS \cite{bengoetxea-et-al-2024}, RefutES\footnote{https://hf.co/datasets/SINAI/RefutES},
TruthfulQA in Basque, Catalan, Galician and Spanish \cite{figueras2025truthknowslanguageevaluating},
VeritasQA \cite{aula-blasco-etal-2025-veritasqa},
PAES Chile \cite{latam-gpt_2025},
meta4xnli \cite{sanchezbayona2024meta4xnli}, MedExpQA \cite{alonso2024medexpqa},
%, Extractive Explanatory QA \cite{goenaga2023explanatory}, Multilingual-BioASQ-6B \cite{garcia-ferrero-etal-2024-medmt5},
Catalonia Independence Corpus (CIC) in Catalan and Spanish \cite{Zotova2021SemiautomaticGO}, HAHA humor detection and analysis in Spanish~\cite{chiruzzo2021overview}, QuALES for question-answering in Spanish in the COVID-19 domain~\cite{rosa2022overview},
AmericasNLP-MT \cite{mager2021findings},
AmericasNLI \cite{ebrahimi2021americasnli},
TraduLATAM, and VocesOriginarias evaluating indigenous languages.

\subsection*{Evaluation dataset details}
The Tables \ref{tab:es_corpora} (Spanish), \ref{tab:ca_corpora} (Catalan), \ref{tab:eu_corpora} (Basque), and \ref{tab:gl_corpora} (Galician) list these datasets, providing additional information about their task type, domain, and origin.
We run the evaluations using our fork of the LM Evaluation Harness\footnote{\url{https://github.com/somosnlp/lm-evaluation-harness}}, synced with the main repository on commit 6ccd520f3fb2b5d74c6f14c05f9d189521424719. The tables mentioned also include details about the evaluation configuration, providing the Harness task ID, metric, and number of shots.

% OPTIONAL STYLE: Order datasets per task and then per domain

\begin{table*}
  \centering
  \scriptsize
\begin{tabular}{lllllll}
    \hline
    \textbf{Dataset} & \textbf{Task} & \textbf{Metric} & \textbf{Domain} & \textbf{Origin} & \textbf{\#Examples} & \textbf{\#Shots} \\
    \hline
    AQuAS & Abstractive QA, Long Form QA & sas\_encoder & Miscellaneous & Original & 87 & 1 \\
    Belebele Spa & Reading Comprehension & acc & Miscellaneous & Human translation & 900 & 2 \\
    ClinDiagnosES & Long Form QA & sas\_encoder & Clinical & Original & 62 & 0 \\
    ClinTreatES & Long Form QA & sas\_encoder & Clinical & Original & 62 & 0 \\
    COPA\_es & Commonsense Reasoning & acc & Lang. prof., Misc. & Human translation & 500 & 3 \\
    Crows Pairs Spanish & Stereotype Detection & pct\_stereotype & Ethics, Hate speech & Original & 1509 & 0 \\
    EsCoLA & Linguistic Acceptability & mcc & Language proficiency & Original & 1060 & 2 \\
    Fake News ES & Fake News Detection & acc & Press & Original & 572 & 2 \\
    HumorQA & Humor Classification & acc & Language proficiency & Original & 51 & 0 \\
    MGSM\_es & Math Reasoning & exact\_match & Math & Human translation & 250 & 2 \\
    NoticIA & Summarization & rouge1 & Lang. prof., Press & Original & 100 & 0 \\
    OffendES & Hate Speech Detection & acc & Hate speech & Original & 13600 & 2 \\
    OpenBookQA\_es & Multiple Choice QA & acc & General knowledge & Human translation & 500 & 0 \\
    PAWS-X\_es & Paraphrasing & acc & Lang. prof., Misc. & Original & 2000 & 3 \\
    RagQuAS & Abstractive QA, Long Form QA & sas\_encoder & Miscellaneous & Original & 201 & 1 \\
    SpaLawEx & Multiple Choice QA & acc & Legal & Original & 119 & 0 \\
    TELEIA & Multiple Choice QA & acc & Gen. knowl., Lang. prof. & Original & 100 & 2 \\
    WNLI ES & Natural Language Inference & acc & Lang. prof., Misc. & Human translation & 146 & 2 \\
    XL-Sum\_es & Summarization & bleu & Press & Original & 4763 & 1 \\
    XNLI\_es & Natural Language Inference & acc & Miscellaneous & Original & 5010 & 3 \\
    XQuAD\_es & Extractive QA & f1 & Miscellaneous & Original & 1190 & 2 \\
    xStoryCloze\_es & Commonsense Reasoning & acc & Miscellaneous & Human translation & 1510 & 0 \\
    \hline
\end{tabular}
  \caption{Details of the evaluation datasets in Spanish (ES).}
  \label{tab:es_corpora}
\end{table*}

\begin{table*}
  \centering
  \scriptsize
\begin{tabular}{lllllll}
    \hline
    \textbf{Dataset} & \textbf{Task} & \textbf{Metric} & \textbf{Domain} & \textbf{Origin} & \textbf{\#Examples} & \textbf{\#Shots} \\
    \hline
    ARC\_ca & Multiple Choice QA & acc & Science & Human translation & 869 & 2 \\
    Belebele Cat & Reading Comprehension & acc & Miscellaneous & Human translation & 900 & 2 \\
    caBREU & Summarization & bleu & Press & Original & 301 & 1 \\
    CatalanQA & Extractive QA & f1 & Miscellaneous & Original & 2135 & 2 \\
    CatCoLA & Linguistic Acceptability & mcc & Language proficiency & Original & 1020 & 2 \\
    COPA\_ca & Commonsense Reasoning & acc & Lang. prof., Misc. & Human translation & 500 & 3 \\
    CoQCat & Extractive QA & f1 & Miscellaneous & Original & 8986 & 1 \\
    MGSM\_ca & Math Reasoning & exact\_match & Math & Human translation & 250 & 2 \\
    OpenBookQA\_ca & Multiple Choice QA & acc & General knowledge & Human translation & 500 & 0 \\
    Parafraseja & Paraphrasing & acc & Language proficiency & Original & 21984 & 3 \\
    PAWS\_ca & Paraphrasing & acc & Lang. prof., Misc. & Human translation & 2000 & 3 \\
    PIQA\_ca & Multiple Choice QA & acc & General knowledge & Human translation & 1838 & 2 \\
    SIQA\_ca & Multiple Choice QA & acc & General knowledge & Human translation & 1954 & 2 \\
    TE-ca & Natural Language Inference & acc & Lang. prof., Misc. & Original & 2117 & 3 \\
    WNLI\_ca & Natural Language Inference & acc & Lang. prof., Misc. & Human translation & 146 & 2 \\
    XNLI\_ca & Natural Language Inference & acc & Lang. prof., Misc. & Human translation & 5010 & 3 \\
    XQuAD\_ca & Extractive QA & f1 & Miscellaneous & Human translation & 1190 & 2 \\
    xStoryCloze\_ca & Commonsense Reasoning & acc & Miscellaneous & Human translation & 1510 & 0 \\
    \hline
\end{tabular}
  \caption{Details of the evaluation datasets in Catalan (CA).}
  \label{tab:ca_corpora}
\end{table*}

\begin{table*}
  \centering
  \scriptsize
\begin{tabular}{lllllll}
    \hline
    \textbf{Dataset} & \textbf{Task} & \textbf{Metric} & \textbf{Domain} & \textbf{Origin} & \textbf{\#Examples} & \textbf{\#Shots} \\
    \hline
    BEC2016eu & Sentiment Analysis & f1 & Politics, Twitter & Original & 1302 & 3 \\
    Belebele Eus & Reading Comprehension & acc & Miscellaneous & Human translation & 900 & 2 \\
    BertaQA & Multiple Choice QA & acc & Cultural Knowledge & Original & 4760 & 3 \\
    BHTCv2 & Topic Classification & f1 & Press & Original & 1854 & 2 \\
    EpecKorrefBin & Natural Language Inference & acc & Press & Original & 587 & 3 \\
    EusExams & Multiple Choice QA & acc & Miscellaneous & Original & 16000 & 4 \\
    EusProficiency & Multiple Choice QA & acc & Language proficiency & Original & 5169 & 4 \\
    EusReading & Reading Comprehension & acc & Miscellaneous & Original & 352 & 1 \\
    EusTrivia & Multiple Choice QA & acc & General knowledge & Original & 1715 & 4 \\
    MGSM\_eu & Math Reasoning & exact\_match & Math & Human translation & 250 & 2 \\
    QNLIeu & Natural Language Inference & acc & Miscellaneous & Original & 238 & 2 \\
    VaxxStance & Stance Detection & f1 & Politics, Twitter & Original & 312 & 3 \\
    WiCeu & Natural Language Inference & acc & Language proficiency & Original & 1400 & 2 \\
    WNLI\_eu & Natural Language Inference & acc & Lang. prof., Misc. & Human translation & 146 & 2 \\
    XCOPA\_eu & Commonsense Reasoning & acc & Lang. prof., Misc. & Human translation & 500 & 3 \\
    XNLI\_eu & Natural Language Inference & acc & Lang. prof., Misc. & Reviewed MT & 5010 & 3 \\
    xStoryCloze\_eu & Commonsense Reasoning & acc & Miscellaneous & Human translation & 1510 & 0 \\
    \hline
\end{tabular}
  \caption{Details for evaluation datasets in Basque (EU).}
  \label{tab:eu_corpora}
\end{table*}

\begin{table*}
    \scriptsize
    \centering
\begin{tabular}{lllllll}
    \hline
    \textbf{Dataset} & \textbf{Task} & \textbf{Metric} & \textbf{Domain} & \textbf{Origin} & \textbf{\#Examples} & \textbf{\#Shots} \\
    \hline
    Belebele Glg & Reading Comprehension & acc & Miscellaneous & Reviewed MT & 900 & 2 \\
    GalCoLA & Linguistic Acceptability & mcc & Language proficiency & Original & 1710 & 2 \\
    MGSM\_gl & Math Reasoning & exact\_match & Math & Reviewed MT & 250 & 2 \\
    OpenBookQA\_gl & Multiple Choice QA & acc & General knowledge & Reviewed MT & 500 & 0 \\
    ParafrasesGL & Paraphrasing & acc & Language proficiency & Original & 294 & 3 \\
    PAWS\_gl & Paraphrasing & acc & Lang. prof., Misc. & Reviewed MT & 2000 & 3 \\
    SummarizationGL & Summarization & bleu & Press & Original & 8080 & 1 \\
    XNLI\_gl & Natural Language Inference & acc & Lang. prof., Misc. & Reviewed MT & 5010 & 3 \\
    xStoryCloze\_gl & Commonsense Reasoning & acc & Miscellaneous & Human translation & 1510 & 0 \\
    \hline
\end{tabular}
  \caption{Details for evaluation datasets in Galician (GL).}
  \label{tab:gl_corpora}
\end{table*}

\clearpage
\section{Frontend Detailed Description}
\label{sec:appendix_frontend}

The implementation of \laleaderboard is based on the HuggingFace leaderboard template. \footnote{\url{https://hf.co/spaces/demo-leaderboard-backend/leaderboard}} The frontend is developed using Gradio \cite{abid2019gradio} and divided into four tabs:
\begin{itemize}
    \item The landing tab, named "La Leaderboard", is divided into five sub-tabs, each containing tables with all the evaluated models and their corresponding average results. These sub-tabs include overall and language-specific results for Spanish, Catalan, Basque, and Galician. The results are aggregated by averaging the scores across all tasks for each language.
    \item For transparency and reproducibility purposes, the second tab, "Info", includes the command we use to evaluate the models and also the normalization formula. In the acknowledgements section, we list the institutions and every person who contributed to the project.
    \item The next tab describes all the "Tasks" included in \laleaderboard.
    \item Finally, there is a tab where everyone can submit their model for evaluation.
\end{itemize}

The text of the information and submission tabs is available both in English and Spanish to bring the tool closer to the community. \\

In the footer, we can find the citation information for the software, all the included datasets, and the evaluation suite. Below are the fourteen logos from all the collaborating institutions.
The entities in the acknowledgements are ordered chronologically by the date they joined the project to thank early adopters, whereas the logos in the footer are ordered by the number of datasets donated.

%\section{Correlation with LLM Arena}
% https://lmarena.ai/leaderboard/text/spanish
% FUTURE WORK: To validate \laleaderboard’s real-world utility, we compare its results against human preferences by correlating them with those from the LLM Arena. We find that...
% La idea está bien pero solo hay 4 modelos en ambas leaderboards

% \section{Dataset samples}
% FUTURE WORK: Add a table at this point with some samples from the tasks to help the reader understand what these tasks are about precisely

\section{Models Evaluated}
\label{sec:appendix_models_evaluated}

Table~\ref{tab:models_evaluated} details the 50 models evaluated, including the following families:
Aitana\footnote{\url{https://hf.co/gplsi/Aitana-6.3B}},
BERTIN \cite{BERTIN}, 
Carballo \cite{gamallo2024openllmsgalician},
Gromenauer\footnote{\url{https://hf.co/bertin-project}}, 
FLOR \cite{da-dalt-etal-2024-flor}, 
LeniaChat\footnote{\url{https://hf.co/LenguajeNaturalAI/leniachat-gemma-2b-v0}},
RigoChat \cite{iic2025rigochat},
Salamandra\footnote{\url{https://hf.co/collections/BSC-LT/salamandra-66fc171485944df79469043a}}, 
Occiglot\footnote{\url{https://hf.co/collections/occiglot/occiglot-eu5-7b-v01-65dbed502a6348b052695e01}}, 
EuroLLM \cite{martins2024eurollmmultilinguallanguagemodels},
Aya \cite{dang2024ayaexpansecombiningresearch},
DeepSeek \cite{deepseekai2025deepseekr1incentivizingreasoningcapability}, 
Gemma \cite{team2024gemma2}, 
Llama \cite{llama3},
Mistral \cite{mistral7b}, 
Phi \cite{phi15}, 
SmolLM \cite{allal2025smollm2smolgoesbig},
and
Qwen \cite{qwen25}.

\begin{table*}[!ht]
\centering
\small
\begin{tabular}{llll}
\toprule
\textbf{Family} & \textbf{Model ID} & \textbf{Model Type} & \textbf{Size (B)} \\ % & \textbf{Languages} \\
\midrule
Aitana & gplsi/Aitana-6.3B & pretrained & 6.25 \\ % & Val 100\% \\
BERTIN & bertin-project/bertin-gpt-j-6B & pretrained & 6.06 \\ % & Spa 100\% \\
BERTIN & bertin-project/Gromenauer-7B & pretrained & 7.24 \\ % & Spa 100\%\\
BERTIN & bertin-project/Gromenauer-7B-Instruct & instruction-tuned & 7.24 \\ % & Spa 100\% \\
Carballo & proxectonos/Carballo-bloom-1.3B & pretrained & 1.31 \\ % & Gal 100\% \\
FLOR & projecte-aina/FLOR-1.3B & pretrained & 1.31 \\ % & Spa/Cat 41\%\\
FLOR & projecte-aina/FLOR-1.3B-Instructed & instruction-tuned & 1.31 \\ % & Spa/Cat 41\%\\
FLOR & projecte-aina/FLOR-6.3B & pretrained & 6.25 \\ % & Spa/Cat 41\%  \\
FLOR & projecte-aina/FLOR-6.3B-Instructed & instruction-tuned & 6.25 \\ % & Spa/Cat 41\% \\
Latxa & HiTZ/latxa-7b-v1.2 & pretrained & 7.00 \\ % & Basque \\
LeniaChat & LenguajeNaturalAI/leniachat-gemma-2b-v0 & instruction-tuned & 2.51 \\ % & Spa 100\% \\
LeniaChat & LenguajeNaturalAI/leniachat-qwen2-1.5B-v0 & instruction-tuned & 1.54 \\ % & Spa 100\%\\
RigoChat & IIC/RigoChat-7b-v2 & instruction-tuned & 7.62 \\ % & Spa 100\% \\
Salamandra & BSC-LT/salamandra-2b & pretrained & 2.25 \\ % & Spa 16\% \\
Salamandra & BSC-LT/salamandra-2b-instruct & instruction-tuned & 2.25 \\ % & Spa 16\% \\
Salamandra & BSC-LT/salamandra-7b & pretrained & 7.77 \\ % & Spa 16\%  \\
Salamandra & BSC-LT/salamandra-7b-instruct & instruction-tuned & 7.77 \\ % & Spa 16\%  \\
\midrule
EuroLLM & utter-project/EuroLLM-1.7B & pretrained & 1.70 \\ % & Multi (n/a)\\
EuroLLM & utter-project/EuroLLM-1.7B-Instruct & instruction-tuned & 1.70 \\ % & Multi (n/a)\\
EuroLLM & utter-project/EuroLLM-9B & pretrained & 9.15 \\ % & Multi (n/a)\\
EuroLLM & utter-project/EuroLLM-9B-Instruct & instruction-tuned & 9.15 \\ % & Multi (n/a)\\
Occiglot & occiglot/occiglot-7b-es-en & pretrained & 7.24 \\ % & Spa 52\% \\
Occiglot & occiglot/occiglot-7b-es-en-instruct & instruction-tuned & 7.24 \\ %  & Spa 52\% \\
Occiglot & occiglot/occiglot-7b-eu5 & pretrained & 7.24 \\ % & Spa 20\% \\
Occiglot & occiglot/occiglot-7b-eu5-instruct & instruction-tuned & 7.24 \\ % & Spa 20\% \\
% Teuken & openGPT-X/Teuken-7B-instruct-research-v0.4 & instruction-tuned & 7.45 \\
\midrule
Aya & CohereForAI/aya-expanse-8b & pretrained & 8.03 \\ % & Multi (n/a)\\
% Aya & CohereForAI/aya-expanse-32b & instruction-tuned & 32.30 \\
DeepSeek & deepseek-ai/DeepSeek-R1-Distill-Qwen-1.5B & instruction-tuned & 1.78 \\ % & Multi (n/a)\\
DeepSeek & deepseek-ai/DeepSeek-R1-Distill-Qwen-7B & instruction-tuned & 7.62 \\ % & Multi (n/a)\\
% DeepSeek & deepseek-ai/DeepSeek-R1-Distill-Llama-8B & instruction-tuned & 8.03 \\
DeepSeek & unsloth/DeepSeek-R1-Distill-Qwen-14B-bnb-4bit & instruction-tuned & 14.8 (8.37) \\ % & Multi (n/a)\\
Gemma & google/gemma-2-2b & pretrained & 2.61 \\ % & Multi (n/a)\\
Gemma & google/gemma-2-2b-it & instruction-tuned & 2.61 \\ % & Multi (n/a)\\
Gemma & google/gemma-2-9b & pretrained & 9.24 \\ % & Multi (n/a)\\
Gemma & google/gemma-2-9b-it & instruction-tuned & 9.24 \\ % & Multi (n/a)\\
Llama & meta-llama/Llama-3.2-1B & pretrained & 1.24 \\ % & Multi (n/a)\\
Llama & meta-llama/Llama-3.2-1B-Instruct & instruction-tuned & 1.24 \\ % & Multi (n/a)\\
% Llama & meta-llama/Llama-3.2-3B & pretrained & 3.21 \\
% Llama & meta-llama/Llama-3.2-3B-Instruct & instruction-tuned & 3.21 \\
Llama & meta-llama/Meta-Llama-3.1-8B & pretrained & 8.03 \\ % & Multi (n/a)\\
Llama & meta-llama/Meta-Llama-3.1-8B-Instruct & instruction-tuned & 8.03 \\ % & Multi (n/a)\\
% Llama & TheBloke/Llama-2-13B-chat-GPTQ & instruction-tuned & 2.03 \\
% Llama & TheBloke/Llama-2-70B-GPTQ & pretrained & 9.10 \\
% Llama & TheBloke/CodeLlama-70B-Instruct-AWQ & instruction-tuned & 9.68 \\
Mistral & mistralai/Mistral-7B-Instruct-v0.3 & instruction-tuned & 7.25 \\ % & Multi (n/a)\\
Mistral & mistralai/Mistral-7B-v0.3 & pretrained & 7.25 \\ % & Multi (n/a)\\
% Mistral & TheBloke/Mixtral-8x7B-v0.1-GPTQ & pretrained & 6.09 \\
Phi & microsoft/phi-1\_5 & pretrained & 1.42 \\ % & Multi (n/a)\\
% Phi & microsoft/Phi-3-small-128k-instruct & instruction-tuned & 7.39 \\
% Phi & microsoft/Phi-3.5-mini-instruct & instruction-tuned & 3.82 \\
SmolLM & HuggingFaceTB/SmolLM2-1.7B & pretrained & 1.71 \\ % & Multi (n/a)\\
SmolLM & HuggingFaceTB/SmolLM2-1.7B-Instruct & instruction-tuned & 1.71 \\ % & Multi (n/a)\\
Qwen & Qwen/Qwen2.5-1.5B & pretrained & 1.54 \\ % & Multi (n/a)\\
Qwen & Qwen/Qwen2.5-1.5B-Instruct & instruction-tuned & 1.54 \\ % & Multi (n/a)\\
% Qwen & Qwen/Qwen2.5-3B & pretrained & 3.09 \\
% Qwen & Qwen/Qwen2.5-3B-Instruct & instruction-tuned & 3.09 \\
Qwen & Qwen/Qwen2.5-7B & pretrained & 7.62 \\ % & Multi (n/a)\\
Qwen & Qwen/Qwen2.5-7B-Instruct & instruction-tuned & 7.62 \\ % & Multi (n/a)\\
Qwen & Qwen/Qwen2.5-14B-Instruct-GPTQ-Int8 & instruction-tuned & 14.80 (4.99) \\ % & Multi (n/a)\\
Qwen & Qwen/Qwen2.5-32B-Instruct-GPTQ-Int4 & instruction-tuned & 32.80 (5.74) \\ % & Multi (n/a)\\
\bottomrule
\end{tabular}
\caption{Models evaluated in \laleaderboard as of February 2025. The table is divided into sections starting at the top with models optimized for the languages of Spain, followed by models created by European projects, and finally models trained by international large technological companies and startups. The size is specified in billions of parameters, as appears in the SafeTensors information of the corresponding Hugging Face model page. For quantized models, the SafeTensors equivalent size of the model is added in parenthesis after the size of the base model.}
\label{tab:models_evaluated}
\end{table*}
% FUTURE WORK: Give the weights in Gb and include the precision, mention how the weights are computed, plus explain why you kept the safetensors equivalent to account for precision differences
% Safetensors: https://github.com/huggingface/huggingface.js/blob/main/packages/hub/src/lib/parse-safetensors-metadata.ts#L264

\begin{table*}[!ht]
\centering
\small % \scriptsize
\scalebox{0.95}{
\begin{tabular}{llllll}
\toprule
\textbf{Model Name} 
  & \makecell{\textbf{Training}\\\textbf{Tokens (B)}} 
  & \makecell{\textbf{Languages}\\\textbf{Included}} 
  & \makecell{\textbf{Tokens (B)}\\\textbf{per Language}} 
  & \textbf{Hardware} 
  & \makecell{\textbf{Training}\\\textbf{Time (h)}} \\
\midrule
Aitana-6.3B & 2.6 & CA & 2.6 & 4 A100 & 80 \\ % 2 epochs over the 1.3 data
bertin-gpt-j-6B & 65 & ES & 65 & 1 v3-8 & 4320 \\ % 6 months
Gromenauer-7B & 14.5 & ES & 14.5 & - & - \\
Gromenauer-7B-Instruct & 14.5+ & ES, CA & 14.5+, - & - & - \\
Carballo-bloom-1.3B & 2.1 & GL & 2.1 & 5 A100 & - \\
FLOR-1.3B & 26 & ES, CA & 10.8, 10.9 & Cerebras CS-2 & - \\
FLOR-1.3B-Instructed & 26+ & ES, CA & 10.8, 10.9+ & Cerebras CS-2 & -\\
FLOR-6.3B & 140 & ES, CA & 46.7, 46.8 & 16 Cereb. CS-2 & 60\\
FLOR-6.3B-Instructed & 140+ & ES, CA & 46.7, 46.8+ & 16 Cereb. CS-2& 60+\\
latxa-7b-v1.2 & 4.2 & EU & 4.2 & 4 A100 & 953\\ % 952.53
leniachat-gemma-2b-v0 & - & ES & - & - & - \\
leniachat-qwen2-1.5B-v0 & - & ES & - & - & - \\
RigoChat-7b-v2 & - & ES & - & 1 A100 & 8.5 \\
salamandra-2b & 12875000 & ES, CA, GL, EU & 2075k, 253k, 39k, 30k & 256 Hopper  & 221k\\ % 256 * 36 days * 24h
salamandra-2b-instruct & 12875000+ & ES, CA, GL, EU & 2075k, 253k, 39k, 30k+ & 256 Hopper & 221k+12\\
salamandra-7b & 12875000 & ES, CA, GL, EU & 2075k, 253k, 39k, 30k & 512 Hopper & 602k\\ % 512 * 49 days * 24h
salamandra-7b-instruct & 12875000+ & ES, CA, GL, EU & 2075k, 253k, 39k, 30k+ & 512 Hopper & 602k+16\\
\midrule
EuroLLM-1.7B & 4000 & ES, CA, GL & 240, 12, 4 & 256 H100 & - \\
EuroLLM-1.7B-Instruct & 4000+ & ES, CA & 240, 12, 4+ & 256 H100 & - \\
EuroLLM-9B & 4000 & ES, CA, GL & 240, 12, 4 & 400 H100 & - \\
EuroLLM-9B-Instruct & 4000+ & ES, CA & 240, 12, 4+ & 400 H100 & - \\ % IT trained on https://huggingface.co/datasets/utter-project/EuroBlocks-SFT-Synthetic-1124
occiglot-7b-es-en & 112 & ES & 58.2 & 128 A100 & - \\
occiglot-7b-es-en-instruct & 112+0.16 & ES & 58.2+0.08 & +8 H100 & - \\
occiglot-7b-eu5 & 293 & ES & 58.6 & 128 A100 & - \\
occiglot-7b-eu5-instruct & 293+0.40 & ES & 58.6+0.08 & +8 H100 & - \\
\midrule
aya-expanse-8b & - & ES & - & - & - \\
DeepSeek-R1-Dist-Qwen-1.5B & - & - & - & - & - \\
DeepSeek-R1-Dist-Qwen-7B & - & - & - & - & - \\
DeepSeek-R1-Dist-Qwen-14B-4bit & - & - & - & - & - \\
gemma-2-2b & 2000 & - & - & v5p & - \\
gemma-2-2b-it & 2000+ & - & - & v5p & - \\
gemma-2-9b & 8000 & - & - & v5p & - \\
gemma-2-9b-it & 8000+ & - & - & v5p & - \\
Llama-3.2-1B & 9000 & ES & - & H100 & 370k \\ % H100-80GB
Llama-3.2-1B-Instruct & 9000+ & ES & - & H100 & 370k+ \\
Llama-3.1-8B & 15000 & ES & - & H100 & 1460k \\ % H100-80GB, 8 lang (English, German, French, Italian, Portuguese, Hindi, Spanish, and Thai)
Llama-3.1-8B-Instruct & 15000+ & ES & - & H100 & 1460k+ \\
Mistral-7B-v0.3 & - & - & - & - & - \\
Mistral-7B-Instruct-v0.3 & - & - & - & - & - \\
phi-1\_5 & 150 & none & 0 & 32 A100 & 192 \\ % 8 days * 24
SmolLM2-1.7B & 11000 & none & 0 & 256 H100 & - \\
SmolLM2-1.7B-Instruct & 11000+ & none & 0 & 256 H100 & - \\
Qwen2.5-1.5B & 18000 & ES & - & - & - \\ % https://qwenlm.github.io/blog/qwen2.5, 29 lang (inc. French, Spanish, Portuguese, Italian)
Qwen2.5-1.5B-Instruct & 18000 & ES & - & - & - \\
Qwen2.5-7B & 18000 & ES & - & - & - \\
Qwen2.5-7B-Instruct & 18000 & ES & - & - & - \\
Qwen2.5-14B-Instruct-GPTQ-Int8 & 18000 & ES & - & - & - \\
Qwen2.5-32B-Instruct-GPTQ-Int4 & 18000 & ES & - & - & - \\
\bottomrule
\end{tabular}
} % end de "scalebox"
\caption{Public details about the training corpus and compute resources of the models evaluated on \laleaderboard: total number of tokens in the training data, target languages included in the training data, number of tokens per language, hardware type, and hours of training time.}
\label{tab:model_training_stats}
\end{table*}

\begin{figure*}[t]
  \includegraphics[width=\linewidth]{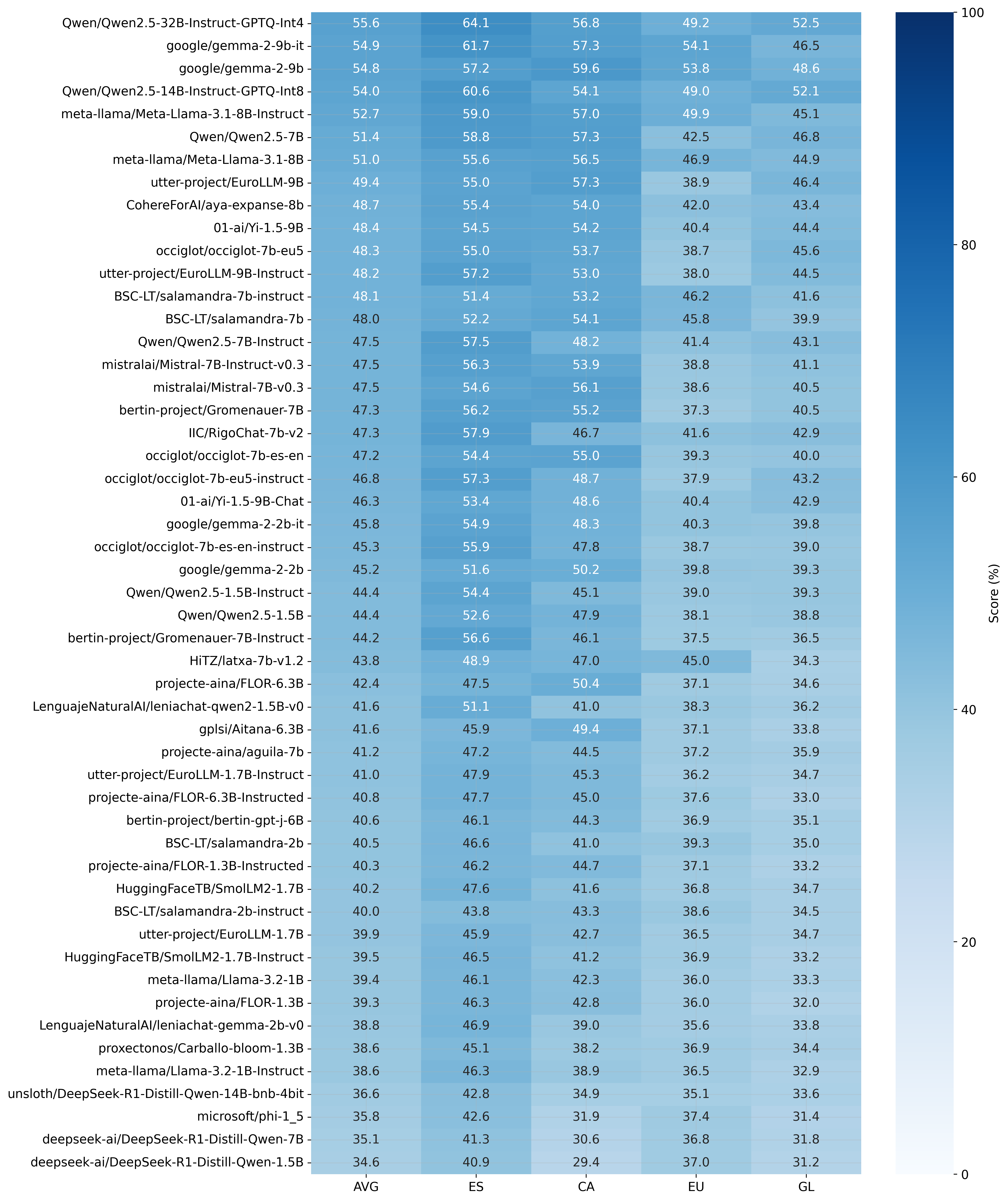}
  \caption{Average results of the first set of models evaluated on \laleaderboard, overall and by language.}
  \label{fig:heatmap_models_avg_scores}
\end{figure*}

%\clearpage

\section{Evaluation Results}
\label{sec:appendix_evaluation_results}

This section detailed visualizations of the evaluation results, comparing models, languages, and tasks, considering metrics such as performance and energy efficiency. 

\begin{figure*}[!h]
  \includegraphics[width=\linewidth]{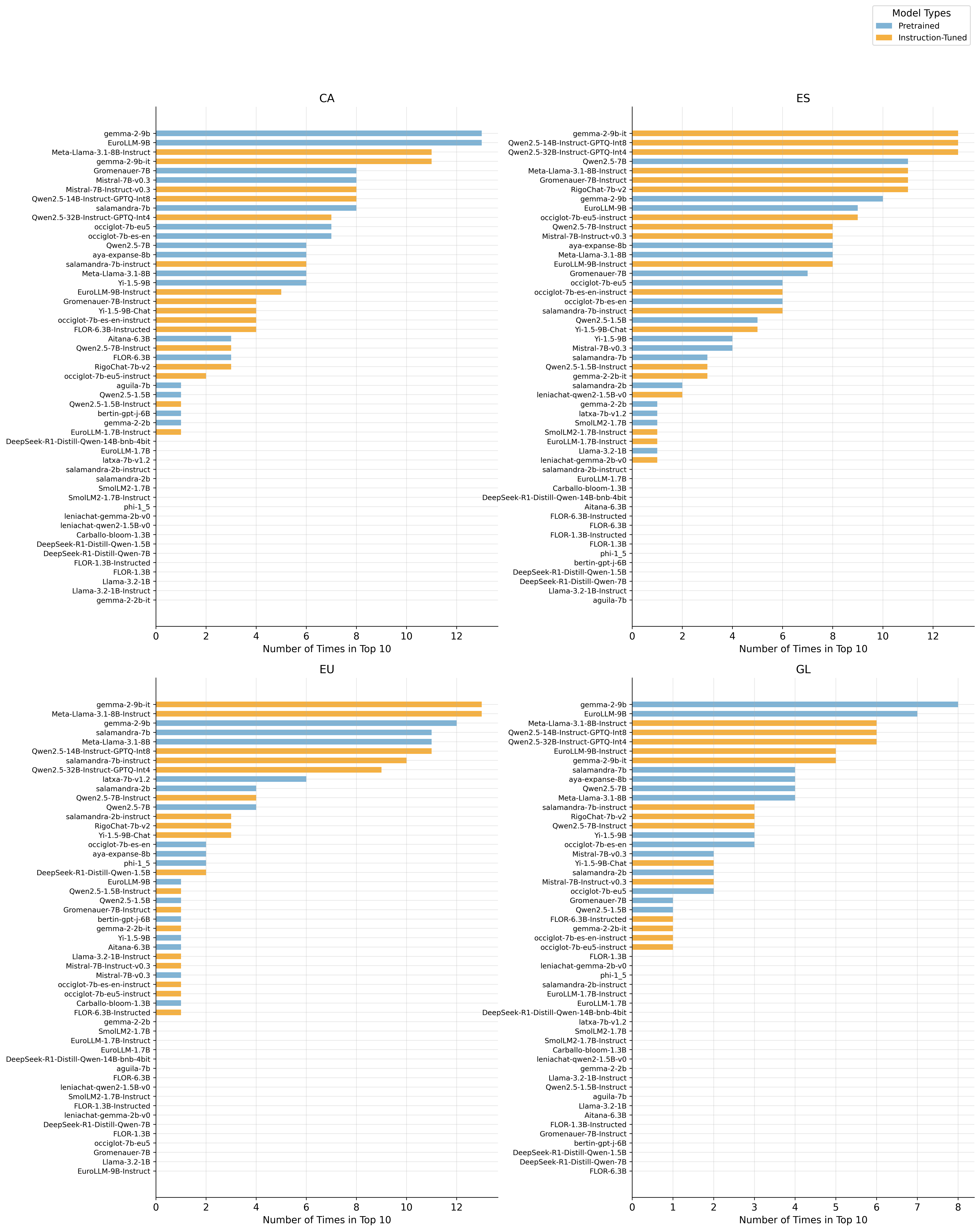}
  \caption{Number of tasks in which a model is among the top 10 models, by language.}
  \label{fig:barplots_top10_frequency_by_language}
\end{figure*}

\begin{figure*}[b]
  \includegraphics[width=\linewidth]{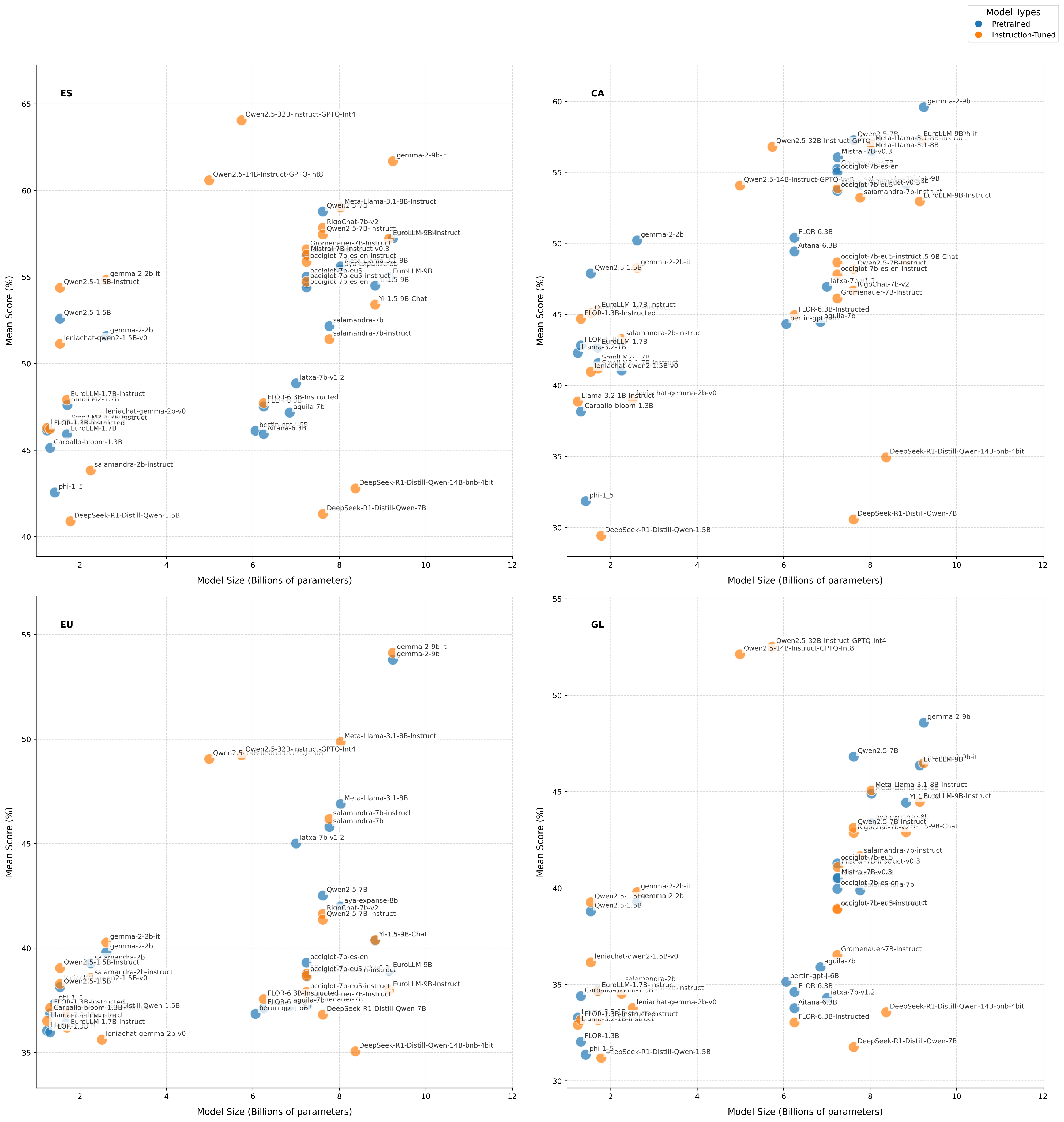}
  \caption{Results of the first set of models evaluated on \laleaderboard organized by language, size, and model type.}
  \label{fig:scatter_plot_performance_vs_size_all_languages}
\end{figure*}

Figure~\ref{fig:heatmap_models_avg_scores} presents the full list of average results of the models evaluated, overall and by language. Moreover, since a very bad or good score in a few tasks can lower or raise the average score for a model and distort the comparison, we show the results in terms of the number of tasks for which a model is in the top 10 in Figure~\ref{fig:barplots_top10_frequency_by_language}. This provides an alternative view of the results, focusing on the number of tasks for which the performance of the model is good.
%It can be seen that the top 5 models are the same as before, but the order changes. Now the Gemma models are in the first two positions, Llama in the third, and the two Qwen models in the last two positions.
%The results per language are presented in Figure~\ref{fig:barplots_top10_frequency_by_language}.
%It can be seen that Gemma models are the best in all four languages, but the top 5 models change significantly, and some language-optimized models are in top positions. For example, EuroLLM-9B is the second in Catalan and Galician, and Salamandra-7B is the fourth in Basque.  
Figure~\ref{fig:scatter_plot_performance_vs_size_all_languages} provides a comprehensive view of all the evaluation results by language, size, and model type.

\begin{figure*}[t]
  \centering
  %=============================%
  % First mini‐figure (left)    %
  %=============================%
  \begin{minipage}[t]{0.45\textwidth}
    \centering
    \includegraphics[width=\linewidth]{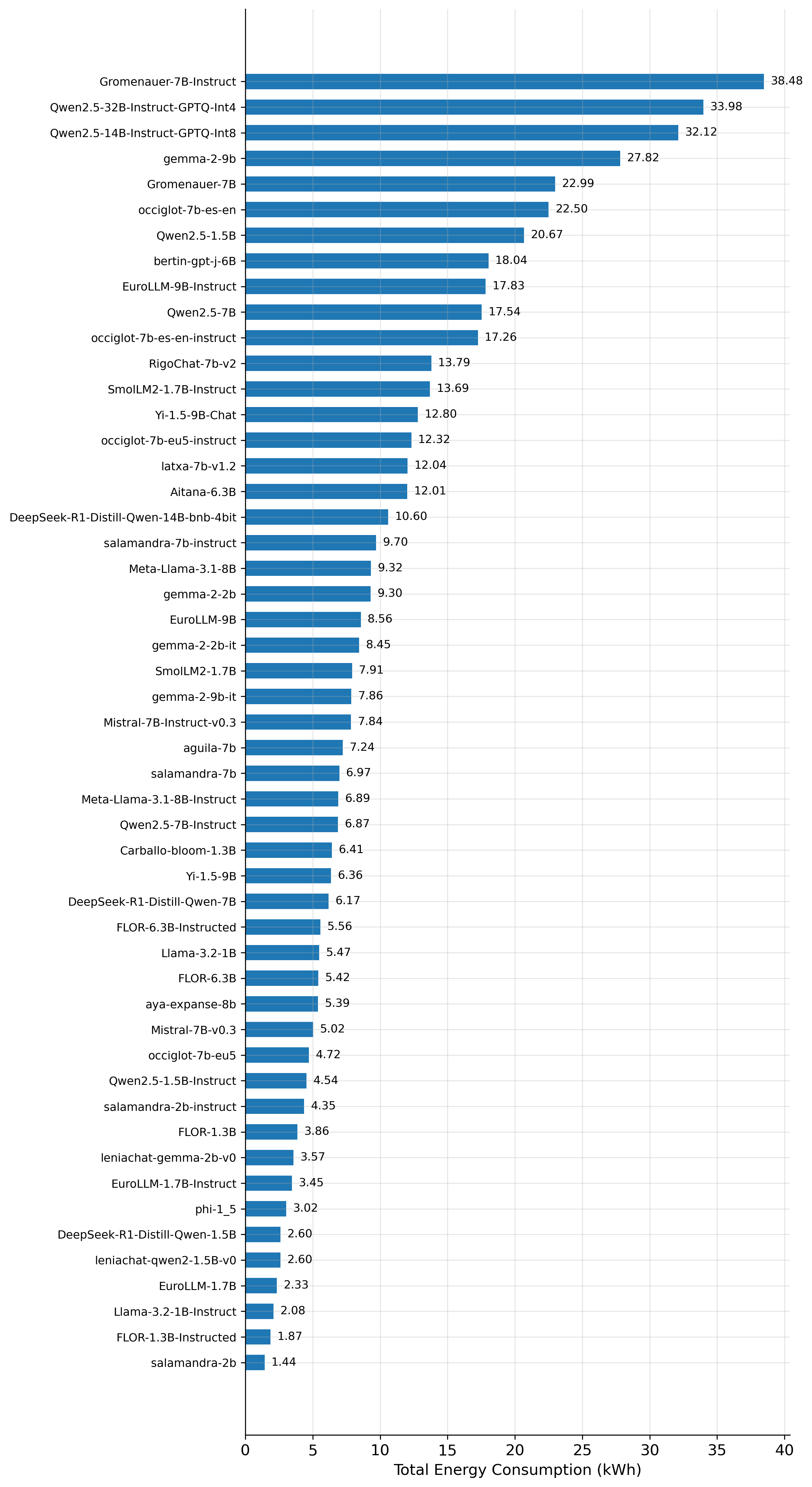}
    \captionof{figure}{Distribution of results of models evaluated on \laleaderboard organized by total energy consumption.}
    \label{fig:carbon_emissions_total}
  \end{minipage}
  \hfill
  %=============================%
  % Second mini‐figure (right)  %
  %=============================%
  \begin{minipage}[t]{0.45\textwidth}
    \centering
    \includegraphics[width=\linewidth]{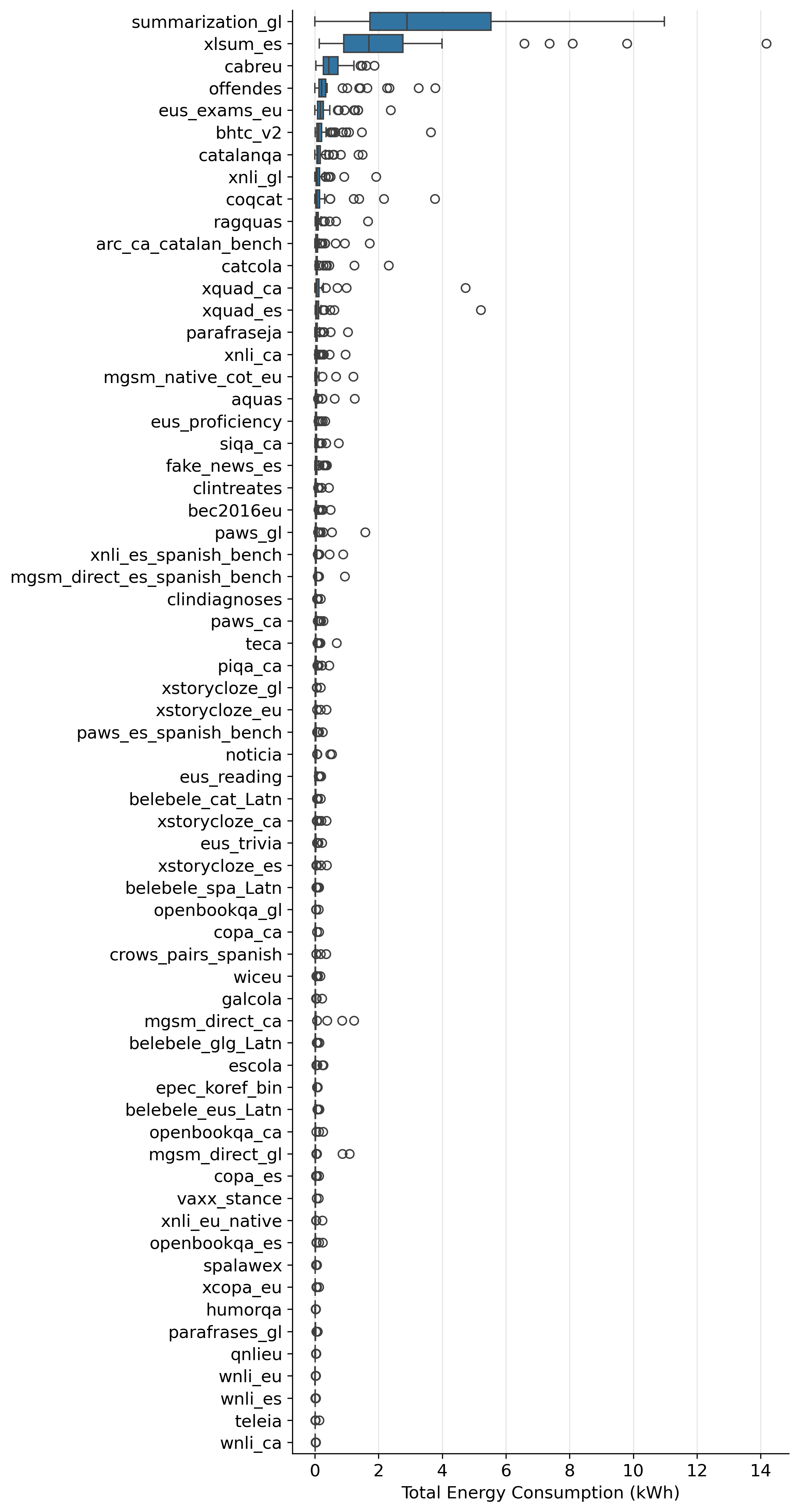}
    \captionof{figure}{Energy consumption for the tasks evaluated on \laleaderboard. The top three tasks (summarization\_gl, xlsum\_es, and cabreu) correspond to text summarization tasks, which require the generation of many tokens. The next tasks correspond to QA tasks with thousands of questions.} % offendes, eus_exams_eu, bhtc_v2, catalanqa, xnli_gl, coqcat % OPTIONAL si comento la de al lado
    \label{fig:carbon_emissions_by_task}
  \end{minipage}
\end{figure*}

\begin{figure*}[h]
  \includegraphics[width=\linewidth]{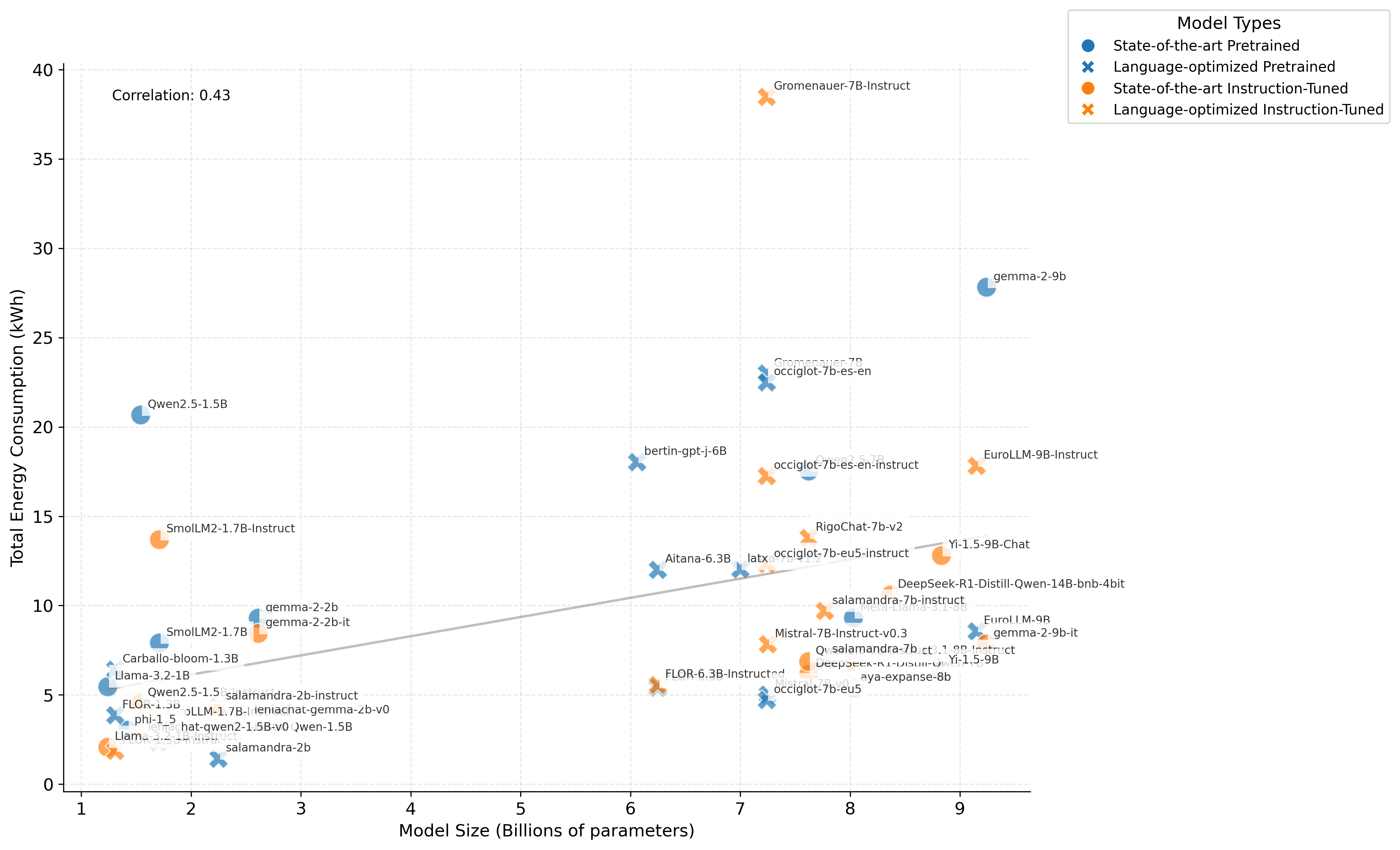}
  \caption{Distribution of results of models evaluated on \laleaderboard comparing energy consumption versus size. The plot shows some correlation between model size and energy consumed, with a Pearson correlation coefficient of 0.43.}
  \label{fig:carbon_emissions_vs_size}
\end{figure*}

\begin{figure*}[h]
  \includegraphics[width=\linewidth]{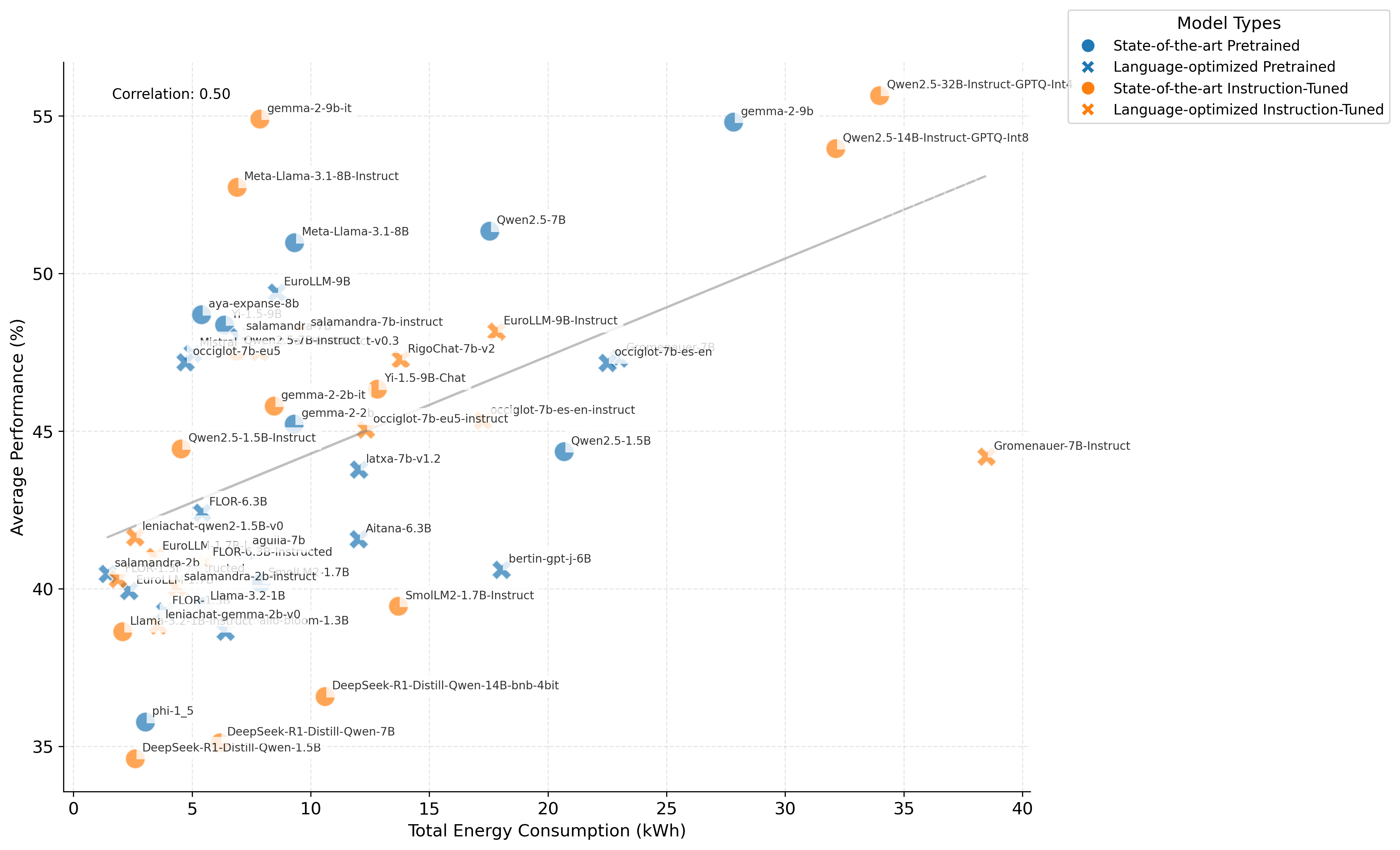}
  \caption{Distribution of results of models evaluated on \laleaderboard comparing energy consumption versus performance. The plot shows a high correlation between model performance and energy consumed, with a Pearson correlation coefficient of 0.50.}
  \label{fig:performance_vs_energy_kwh}
\end{figure*}

Figure~\ref{fig:carbon_emissions_total} represents the total energy consumed by each model. On average, each model consumed 9.25 kWh (median = 6.88, SD = 8.42), showing a wide variety in energy usage.
Figure~\ref{fig:carbon_emissions_by_task} shows which tasks consume more energy, with the text generation tasks on the top of the list.
Figure~\ref{fig:carbon_emissions_vs_size} presents a comparison between model size and energy consumption.
Finally, Figure~\ref{fig:performance_vs_energy_kwh} shows the relation between performance and energy consumption.

\clearpage

%\subsection*{Model efficiency}

%Figure~\ref{fig:carbon_emissions_total} represents the total energy consumed by each model. On average, each model consumed 9.25 kWh (median = 6.88, SD = 8.42), showing a wide variety in energy usage.
%Figure~\ref{fig:carbon_emissions_by_task} shows which tasks consume more energy, with the text generation tasks on the top of the list.
%Figure~\ref{fig:carbon_emissions_vs_size} presents a comparison between model size and energy consumption.
%Finally, Figure~\ref{fig:performance_vs_energy_kwh} shows the relation between performance and energy consumption.

\clearpage
\section{Data Collection Campaign}
\label{sec:appendix_data_collection}

Below are the questions, translated into English, corresponding to the Google Form used in the open data collection campaign.
The asterisk (*) marks mandatory fields.

% REVIEW Add also intro text?
%Campaña de recolección de corpus #Somos600M
% Queremos modelos que entiendan y hablen el español de las 600M personas hispanohablantes. ¿Nos ayudas? Estamos recolectando corpus de diferentes países, registros y dominios. ¡Cuantas más variedades de la lengua, mejor! Son bienvenidos TODXXXX los tipos de corpus: entrenamiento y evaluación, todas las tareas de comprensión (NLU) y generación (NLG), todas las modalidades (texto, audio e imágenes con descripción). También buscamos corpus en otras lenguas habladas en países hispanohablantes (e.g., catalán, quechua).
%Importante: Si quieres compartir varios corpus, rellena este formulario una vez por corpus. Intenta proveer información tan detallada como puedas, si no sabes la respuesta de algún campo escribe NSNC, ¡muchas gracias!
%Más información sobre la campaña en somosnlp.org/donatucorpus. Si tienes cualquier duda manda un email a info@somosnlp.org o contáctanos por Discord. ¡Muchas gracias por apoyarnos en nuestra misión de democratizar el PLN en español!

\begin{enumerate}
\small
    \item Email *
    \item Data source (Select one option) *
    \begin{enumerate}
        \item The dataset is public
        \item Instructions to recreate it are available
        \item The dataset is private but access can be requested on a website
        \item The dataset is currently private, but we want to open it as a donation
        \item The dataset is private, but you should try contacting the organization that created it
    \end{enumerate}
    \item Dataset link * This can be the dataset link, the instructions to recreate it, or the corresponding organization’s website if private.
    \item If your dataset is not uploaded to Hugging Face, would you like us to take care of uploading it? (Select one option)
    \begin{enumerate}
        \item Yes, upload it to the SomosNLP organization
        \item Yes, help me create my own organization and upload it
        \item No, I prefer to create my own organization and upload it myself
    \end{enumerate}
    \item Modality * (Select one option)
    \begin{enumerate}
        \item Text
        \item Audio
        \item Image (e.g., images with descriptions)
    \end{enumerate}
    \item Language(s) * (Select all that apply)
    \begin{enumerate}
        \item Spanish
        \item Other: \_\_\_\_\_
    \end{enumerate}
    \item Country(ies) *  
    Country(ies) of origin of the data and/or the people who annotated it. A region can also be specified if known. The more information, the better.
    \item Tasks * (Select all that apply)
    \begin{enumerate}
        \item Language modeling (unannotated)
        \item Question answering (QA)
        \item Classification
        \item Token classification (e.g., NER, PoS)
        \item Translation
        \item Summarization
        \item Semantic similarity
        \item Multimodal (e.g., text-to-image, audio-to-text)
    \end{enumerate}
    \item Subtask  
    For example, subtasks of "text classification" could be "sentiment analysis" or "hate speech detection."
    \item Domain * (Select all that apply)
    \begin{enumerate}
        \item Legal
        \item Clinical or biomedical
        \item Academic or technical
        \item Literature or music
        \item Social media or forums
        \item News or articles
        \item Dialogues
        \item General
    \end{enumerate}
    \item Number of examples  
    Enter the exact number of examples if known, otherwise provide a range.
    \item License type *
    \begin{enumerate}
        \item Commercial
        \item Non-commercial
    \end{enumerate}
    \item License link
    \item Link to the dataset documentation or any other relevant information: description, annotation and cleaning process, ethical considerations... *
    \item Link to the script/repository on GitHub to download or process the dataset
    \item Thank you very much for your contribution!  
    To publicly acknowledge your contribution, you may share your name and/or affiliation to be displayed on the website. If this is a donation, we will contact you soon—thank you!
    \item Name
    \item Affiliation
    \item How could we improve this campaign? Who would you recommend we contact? Anything else you would like to tell us?
\end{enumerate}

% FUTURE WORK for the indigenous languages:
% If you include the indigenous language benchmarks, please consider the points mentioned here: https://aclanthology.org/2020.coling-main.313/ and https://aclanthology.org/2023.acl-long.268/ (Manuel)

\clearpage

\section{Datasheets}
\label{sec:appendix_datasheets}

We present the datasheets \cite{gebru2021datasheets} corresponding to each of the datasets specifically created for \laleaderboard: AQuAS, ClinDiagnosES, ClinTreatES, HumorQA, RagQuAS, SpaLawEx, TELEIA. Moreover, we propose an adaptation for leaderboards and fill it for \laleaderboard. 

%\url{https://www.microsoft.com/en-us/research/uploads/prod/2019/01/1803.09010.pdf}

% STYLE:
% Remove all NA questions, or aggregate them in a list of "Not applicable questions"
% Only keep the questions, not the examples

\section*{La Leaderboard}

\subsection*{Motivation for Leaderboard Creation}

\paragraph{Why was the leaderboard created?}

\laleaderboard is the first open-source leaderboard to evaluate generative LLMs in languages of Spain and Latin America.
By aiming to address the linguistic and cultural
diversity of the Spanish-speaking community, \laleaderboard aims to set a new standard for multilingual LLM evaluation. Our goal is to encourage the development of models that are not only linguistically competent but also culturally aware, ultimately driving progress in Natural Language Processing (NLP) for the benefit of our whole community.

% \paragraph{What (other) tasks could the dataset be used for?} Are there obvious tasks for which it should not be used?

\paragraph{Who funded the creation of the leaderboard?}

\laleaderboard is an initiative launched by an international open-source community and was promoted by volunteers. The funding of each of the individual datasets donated to \laleaderboard will be disclosed after review.

\subsection*{Leaderboard Composition} 

\paragraph{What are the instances?}

\laleaderboard consists of 66 evaluation datasets. All the evaluation datasets in the leaderboard consist solely of text instances.

\paragraph{Are relationships between instances made explicit in the data} % (e.g., social network links, user/movie ratings, etc.)?

There are no known relationships between instances.

\paragraph{How many instances of each type are there?}

Summing all the instances of the 66 evaluation datasets, the leaderboard consists of 149,782 examples.

%\paragraph{What data does each instance consist of? }“Raw” data (e.g., unprocessed text or images)? Features/attributes? Is there a label/target associated with instances? If the instances are related to people, are subpopulations identified (e.g., by age, gender, etc.) and what is their distribution? 

\paragraph{Is everything included or does the data rely on external resources?} % (e.g., websites, tweets, datasets) If external resources, a) are there guarantees that they will exist, and remain constant, over time; b) is there an official archival version. Are there licenses, fees or rights associated with any of the data?

Everything is included in the datasets.

\paragraph{Are there recommended data splits or evaluation measures?} % (e.g., training, development, testing; accuracy/AUC)

The splits used in \laleaderboard are the corresponding test splits of each dataset.

\subsection*{Data Collection Process}

\paragraph{How was the data collected?}

The datasets were collected through an open data collection campaign.

\paragraph{Who was involved in the data collection process? How were they compensated? } 

Professional researchers from academia and industry. The logo and names of the donators are included in the user interface, and the creators of the datasets are acknowledged in the paper.

\paragraph{Over what time-frame was the data collected?}

During 2024.

%\paragraph{How was the data associated with each instance acquired?} Was the data directly observable (e.g., raw text, movie ratings), reported by subjects (e.g., survey responses), or indirectly inferred/derived from other data (e.g., part of speech tags; model-based guesses for age or language)? If the latter two, were they validated/verified and if so how? 

\paragraph{Does the dataset contain all possible instances?}

The evaluations are launched including all the available test instances for each donated dataset.

\paragraph{If the dataset is a sample, then what is the population?} % What was the sampling strategy (e.g., deterministic, probabilistic with specific sampling probabilities)? Is the sample representative of the larger set (e.g., geographic coverage)? If not, why not (e.g., to cover a more diverse range of instances)? How does this affect possible uses? 
Not applicable.

\paragraph{Is there information missing from the dataset and why?}
No

\paragraph{Are there any known errors, sources of noise, or redundancies in the data?} 
No.

%\subsection*{Data Preprocessing} 

%\paragraph{What preprocessing/cleaning was done?} % (e.g., discretization or bucketing, tokenization, part-of-speech tagging, SIFT feature extraction, removal of instances, processing of missing values, etc.) 

%\paragraph{Was the “raw” data saved in addition to the preprocessed/clean data?}

%\paragraph{Is the preprocessing software available? }

%\paragraph{Does this dataset collection/processing procedure achieve the motivation for creating the dataset stated in the first section of this datasheet? }

\subsection*{Leaderboard Distribution} 

\paragraph{How is the leaderboard distributed?} % (e.g., website, API, etc.; does the data have a DOI; is it archived redundantly?) 
The leaderboard is available in the HuggingFace hub\footnote{\url{https://hf.co/spaces/la-leaderboard/la-leaderboard}}. 

\paragraph{When will the leaderboard be released/first distributed?} % (Is there a canonical paper/reference for this dataset?) 
The leaderboard was released in September 2024.

\paragraph{What license (if any) is it distributed under?} % Are there any copyrights on the data? 
The leaderboard is licensed under "Apache 2.0".

\paragraph{Are there any fees or access/export restrictions? }
There are no fees or restrictions.

\subsection*{Leaderboard Maintenance }

\paragraph{Who is supporting/hosting/maintaining the leaderboard?} % How does one contact the owner/curator/manager of the leaderboard (e.g. email address, or other contact info)? 
The leaderboard is hosted at HuggingFace\footnote{\url{https://hf.co/spaces/la-leaderboard/la-leaderboard}}, and the community can be contacted through the "Discussions" tab in the interface or via email\footnote{\href{mailto:maria.grandury@somosnlp.org}{maria.grandury@somosnlp.org}}.

\paragraph{Will the leaderboard be updated? How often and by whom?} % How will updates/revisions be documented and communicated (e.g., mailing list, GitHub)? Is there an erratum? 
Yes, every time there is a new donation, the maintainer will update the leaderboard and communicate the update on the usual communication channels of the open-source community.

\paragraph{Is there a repository to link to any/all papers/systems that use this leaderboard?}
Yes, all the datasets and tools used by \laleaderboard are referenced in the "Citation" section of the interface\footnote{\url{https://hf.co/spaces/la-leaderboard/la-leaderboard}}.

\subsection*{Legal \& Ethical Considerations }

\paragraph{If the dataset relates to people or was generated by people, were they informed about the data collection?} % (e.g., datasets that collect writing, photos, interactions, transactions, etc.) 
Not applicable.

\paragraph{If it relates to other ethically protected subjects, have appropriate obligations been met? } % (e.g., medical data might include information collected from animals) 
Not applicable.

\paragraph{If it relates to people, were there any ethical review applications/reviews/approvals? }
Not applicable.

\paragraph{If it relates to people, were they told what the dataset would be used for and did they consent? } % What community norms exist for data collected from human communications? If consent was obtained, how? Were the people provided with any mechanism to revoke their consent in the future or for certain uses? 
Not applicable.

\paragraph{If it relates to people, could this dataset expose people to harm or legal action?} % (e.g., financial social or otherwise) What was done to mitigate or reduce the potential for harm? 
Not applicable.

\paragraph{If it relates to people, does it unfairly advantage or disadvantage a particular social group?} % In what ways? How was this mitigated?
Not applicable.

\paragraph{If it relates to people, were they provided with privacy guarantees?} % If so, what guarantees and how are these ensured? 
Not applicable.

\paragraph{Does the dataset comply with the EU General Data Protection Regulation (GDPR)?} % Does it comply with any other standards, such as the US Equal Employment Opportunity Act? 
Yes, it complies with GDPR.

\paragraph{Does the dataset contain information that might be considered sensitive or confidential?} 
No.

\paragraph{Does the dataset contain information that might be considered inappropriate or offensive? }
No.

\newpage
\section*{AQuAS}
The Abstractive Question-Answering in Spanish (AQuAS) dataset \citep{iic2024aquas},
developed by Instituto de Ingeniería del Conocimiento, %a research institute\footnote{Name removed for review.},
is a monolingual Spanish dataset designed for abstractive question-answering. It contains 107 examples covering a diverse range of topics, including finance, insurance, healthcare, music, and law. 
Each example consists of a context passage, a related question, and a human-crafted answer.
The dataset is aimed at evaluating the ability of large language models (LLMs) to generate well-formed, coherent, and informative responses.

\subsection*{Motivation for Dataset Creation }

\paragraph{Why was the dataset created?} % (e.g., were there specific tasks in mind, or a specific gap that needed to be filled?) 

AQuAS was created to provide high-quality examples of pairs of questions and answers with a related context that can be used to evaluate the ability of large language models (LLMs) to generate well-formed, coherent, and informative responses (abstractive question answering). 

\paragraph{What (other) tasks could the dataset be used for?} % Are there obvious tasks for which it should not be used? 
There are no recommended uses for this dataset other than evaluation. 

\paragraph{Who funded the creation of the dataset?} If there is an associated grant, provide the grant number. 
The dataset was created and funded by the research institute.

\subsection*{Dataset Composition} 

\paragraph{What are the instances?} % (that is, examples; e.g., documents, images, people, countries) Are there multiple types of instances? (e.g., movies, users, ratings; people, interactions between them; nodes, edges) 
Each instance is a pair of a question and an answer accompanied by the related context on which the answer has been based and the corresponding topic. 

\paragraph{Are relationships between instances made explicit in the data} % (e.g., social network links, user/movie ratings, etc.)? 
There are no known relationships between instances. 

\paragraph{How many instances of each type are there? }
The dataset consists of 107 examples. 

\paragraph{What data does each instance consist of? } %“Raw” data (e.g., unprocessed text or images)? Features/attributes? Is there a label/target associated with instances? If the instances are related to people, are subpopulations identified (e.g., by age, gender, etc.) and what is their distribution? 

The instances consist of text data and are labelled with the corresponding topic. 

\paragraph{Is everything included or does the data rely on external resources?} % (e.g., websites, tweets, datasets) If external resources, a) are there guarantees that they will exist, and remain constant, over time; b) is there an official archival version. Are there licenses, fees or rights associated with any of the data? 

Everything is included in the dataset.   

\paragraph{Are there recommended data splits or evaluation measures?}

Since the dataset is intended for testing, there is no recommended split. 

\subsection*{Data Collection Process }

\paragraph{How was the data collected?}

The data for the contexts was gathered from different sources on the web using software to crawl those sites. The rest of the dataset (question-answer pairs) was curated and created manually.  

\paragraph{Who was involved in the data collection process? How were they compensated? }

The data was collected by computational linguists and data scientists from a research institute. 

\paragraph{Over what time-frame was the data collected?}

The data was collected during 2023, when the dataset was created. 

\paragraph{How was the data associated with each instance acquired?} % Was the data directly observable (e.g., raw text, movie ratings), reported by subjects (e.g., survey responses), or indirectly inferred/derived from other data (e.g., part of speech tags; model-based guesses for age or language)? If the latter two, were they validated/verified and if so how?

The question-answer pairs were created and revised by computational linguists. 

\paragraph{Does the dataset contain all possible instances?} % Or is it, for instance, a sample (not necessarily random) from a larger set of instances? 

The dataset is composed of selected instances of different datasets created by a research institute. 

\paragraph{If the dataset is a sample, then what is the population?} % What was the sampling strategy (e.g., deterministic, probabilistic with specific sampling probabilities)? Is the sample representative of the larger set (e.g., geographic coverage)? If not, why not (e.g., to cover a more diverse range of instances)? How does this affect possible uses? 

This dataset is a 24,5\% sample of the original complete datasets. The instances were randomly selected from the original datasets. 

\paragraph{Is there information missing from the dataset and why?} % (this does not include intentionally dropped instances; it might include, e.g., redacted text, withheld documents) Is this data missing because it was unavailable?

There is no data missing. 

\paragraph{Are there any known errors, sources of noise, or redundancies in the data?} 

There are no known errors because the revision process ensured the data is as clean and error-free as possible. 

\subsection*{Data Preprocessing} 

\paragraph{What preprocessing/cleaning was done?} % (e.g., discretization or bucketing, tokenization, part-of-speech tagging, SIFT feature extraction, removal of instances, processing of missing values, etc.) 

The text contained in the "context" part of each instance in the dataset has not undergone any preprocessing or changes. There was no need to apply any cleaning to the question-answer pairs because they were created manually by computational linguists following a rigorous methodology and were subjected to revision afterwards. 

\paragraph{Was the “raw” data saved in addition to the preprocessed/clean data?} % (e.g., to support unanticipated future uses) 

No, the text in the dataset is the raw data. 

\paragraph{Is the preprocessing software available? }

No preprocessing software was used. 

\paragraph{Does this dataset collection/processing procedure achieve the motivation for creating the dataset stated in the first section of this datasheet? }

Yes, the collection procedure ensures the dataset is sufficiently varied so it can be used to evaluate a model on a wide range of topics. However, there are some potential limitations in the dataset which might slightly bias the data towards particular topics, because not all topics included have the exact same representation in the dataset, and obviously it was not possible to cover all topics in existence. 

\subsection*{Dataset Distribution} 

\paragraph{How is the dataset distributed?} % (e.g., website, API, etc.; does the data have a DOI; is it archived redundantly?) 

The dataset is available in HuggingFace\footnote{\url{https://hf.co/datasets/IIC/AQuAS}}.

\paragraph{When will the dataset be released/first distributed?} % (Is there a canonical paper/reference for this dataset?) 

The dataset was released in 2024. 

\paragraph{What license (if any) is it distributed under?} % Are there any copyrights on the data? 

The dataset is licensed under \href{https://creativecommons.org/licenses/by-nc-sa/4.0}{CC BY-NC-SA 4.0}. 

\paragraph{Are there any fees or access/export restrictions? }

There are no fees or restrictions. 

\subsection*{Dataset Maintenance }

\paragraph{Who is supporting/hosting/maintaining the dataset?} How does one contact the owner/curator/manager of the dataset? 

The dataset is hosted at HuggingFace, and the research institute can be contacted through email
\href{mailto:contacto.iic@iic.uam.es}{contacto.iic@iic.uam.es}.

\paragraph{Will the dataset be updated? How often and by whom?} How will updates/revisions be documented and communicated? Is there an erratum? 

It is not planned to update the dataset at the moment.   

\paragraph{Is there a repository to link to any/all papers/systems that use this dataset?} 

No.  

\subsection*{Legal \& Ethical Considerations }

\paragraph{If the dataset relates to people or was generated by people, were they informed about the data collection?} % (e.g., datasets that collect writing, photos, interactions, transactions, etc.)

Not applicable. The data was collected from public web sources, and does not contain sensitive personal information. 

\paragraph{If it relates to other ethically protected subjects, have appropriate obligations been met? } %(e.g., medical data might include information collected from animals) 

Not applicable.    

\paragraph{If it relates to people, were there any ethical review applications/reviews/approvals? } %(e.g. Institutional Review Board applications) 

Not applicable. 

\paragraph{If it relates to people, were they told what the dataset would be used for and did they consent? } % What community norms exist for data collected from human communications? If consent was obtained, how? Were the people provided with any mechanism to revoke their consent in the future or for certain uses? 

Not applicable. 

\paragraph{If it relates to people, could this dataset expose people to harm or legal action?} % (e.g., financial social or otherwise) What was done to mitigate or reduce the potential for harm? 

Not applicable. 

\paragraph{If it relates to people, does it unfairly advantage or disadvantage a particular social group?} % In what ways? How was this mitigated? 

Not applicable. 

\paragraph{If it relates to people, were they provided with privacy guarantees?} % If so, what guarantees and how are these ensured? 

Not applicable. 

\paragraph{Does the dataset comply with the EU General Data Protection Regulation (GDPR)?} % Does it comply with any other standards, such as the US Equal Employment Opportunity Act? 

The dataset complies with GDPR. 

\paragraph{Does the dataset contain information that might be considered sensitive or confidential?} % (e.g., personally identifying information) 

No. 

\paragraph{Does the dataset contain information that might be considered inappropriate or offensive? }

No. 

\newpage
\section*{ClinTreatES}
The ClinTreatES \citep{lenguajenaturalai2024medicalexpertes} %(Cite removed for review)
dataset consists of clinical cases collected directly from doctors in various medical specialties (cardiology, traumatology, emergency, psychiatry, neurology, dermatology, ENT-laryngology, and anaesthesia) across European medical centers. It was developed through a joint collaboration between LenguajeNatural.AI % an NLP startup\footnote{Name removed for review.}
and healthcare professionals. The dataset is intended for evaluating the ability of large language models (LLMs) to generate effective treatment plans based on provided clinical cases and diagnoses.

\subsection*{Motivation for Dataset Creation}
\paragraph{\textbf{Why was the dataset created?}}
ClinTreatES was created to evaluate LLMs’ capability to design appropriate treatments from real clinical cases and their corresponding diagnoses.
\paragraph{\textbf{What (other) tasks could the dataset be used for?}}
In addition to treatment planning, the dataset may be used to study medical reasoning and decision-making; however, it is not recommended for diagnostic tasks.
\paragraph{\textbf{Who funded the creation of the dataset?}}
The dataset was developed through a collaboration between an NLP startup
and healthcare professionals.

\subsection*{Dataset Composition}
\paragraph{\textbf{What are the instances?}}
Each instance comprises a clinical case description and its associated diagnosis.
\paragraph{\textbf{Are relationships between instances made explicit in the data?}}
No, there are no explicit relationships between instances.
\paragraph{\textbf{How many instances of each type are there?}}
The dataset contains 62 examples.
\paragraph{\textbf{What data does each instance consist of?}}
Each instance includes text data: a clinical case and its corresponding diagnosis, which serves as the basis for generating a treatment plan.
\paragraph{\textbf{Is everything included or does the data rely on external resources?}}
The dataset is self-contained with no reliance on external resources.
\paragraph{\textbf{Are there recommended data splits or evaluation measures?}}
No specific splits are recommended; the dataset is intended primarily for evaluation purposes.

\subsection*{Data Collection Process}
\paragraph{\textbf{How was the data collected?}}
Data was collected directly from healthcare professionals across various specialities in European medical centers.
\paragraph{\textbf{Who was involved in the data collection process?}}
Medical professionals from cardiology, traumatology, emergency medicine, psychiatry, neurology, dermatology, ENT-laryngology, and anesthesia contributed to the dataset.
\paragraph{\textbf{Over what time-frame was the data collected?}}
The data was collected in 2024.
\paragraph{\textbf{How was the data associated with each instance acquired?}}
Clinical cases and their corresponding diagnoses were directly provided by the contributing healthcare professionals.
\paragraph{\textbf{Does the dataset contain all possible instances?}}
It is a curated collection and does not cover every possible clinical case.
\paragraph{\textbf{If the dataset is a sample, then what is the population?}}
The dataset represents a curated sample of clinical cases from European medical centers.
\paragraph{\textbf{Is there information missing from the dataset and why?}}
No, all relevant information is included.
\paragraph{\textbf{Are there any known errors, sources of noise, or redundancies in the data?}}
The data has been carefully curated and reviewed to minimize errors and noise.

\subsection*{Data Preprocessing}
\paragraph{\textbf{What preprocessing/cleaning was done?}}
The clinical texts were formatted according to a standardized template; only minimal preprocessing was applied.
\paragraph{\textbf{Was the “raw” data saved in addition to the preprocessed/clean data?}}
Yes, the dataset contains the original clinical texts as provided by the contributors.
\paragraph{\textbf{Is the preprocessing software available?}}
No specific preprocessing software was used.
\paragraph{\textbf{Does this dataset collection/processing procedure achieve the motivation for creating the dataset stated in the first section of this datasheet?}}
Yes, the collection and curation process ensures the dataset is suitable for evaluating treatment design tasks by LLMs.

\subsection*{Dataset Distribution}
\paragraph{\textbf{How is the dataset distributed?}}
The dataset is available on HuggingFace\footnote{\url{https://hf.co/datasets/LenguajeNaturalAI/ClinTreatES}}.
\paragraph{\textbf{When will the dataset be released/first distributed?}}
The dataset was released in March 2024.
\paragraph{\textbf{What license (if any) is it distributed under?}}
It is distributed under the CC BY-NC-SA 4.0 license.
\paragraph{\textbf{Are there any fees or access/export restrictions?}}
There are no fees or restrictions.

\subsection*{Dataset Maintenance}
\paragraph{\textbf{Who is supporting/hosting/maintaining the dataset?}}
The dataset is hosted on HuggingFace and maintained by the NLP startup.
\paragraph{\textbf{Will the dataset be updated? How often and by whom?}}
No updates are planned at this time.
\paragraph{\textbf{Is there a repository to link to any/all papers/systems that use this dataset?}}
The dataset is available on HuggingFace\footnote{\url{https://hf.co/datasets/LenguajeNaturalAI/ClinTreatES}}.

\subsection*{Legal \& Ethical Considerations}
\paragraph{\textbf{If the dataset relates to people, were they informed about the data collection?}}
The clinical cases were provided by healthcare professionals; any personal details have been removed to ensure anonymity. They were anonymized by the healthcare professionals themselves, before transferring the data to the NLP startup.
\paragraph{\textbf{If it relates to other ethically protected subjects, have appropriate obligations been met?}}
Yes, all obligations have been met and ensured in the data collection process.
\paragraph{\textbf{If it relates to people, were there any ethical review applications/reviews/approvals?}}
Yes, healthcare professionals ensured the ethical review was complete.
\paragraph{\textbf{If it relates to people, were they told what the dataset would be used for and did they consent?}}
Yes, patients were told in advance about the objective of data collection and they provided their consent for this use.
\paragraph{\textbf{If it relates to people, could this dataset expose people to harm or legal action?}}
No, as the data is anonymized by the healthcare professionals.
\paragraph{\textbf{If it relates to people, does it unfairly advantage or disadvantage a particular social group?}}
No.
\paragraph{\textbf{If it relates to people, were they provided with privacy guarantees?}}
Yes, all personal information has been removed by the healthcare professionals.
\paragraph{\textbf{Does the dataset comply with the EU General Data Protection Regulation (GDPR)?}}
Yes, it complies with GDPR.
\paragraph{\textbf{Does the dataset contain information that might be considered sensitive or confidential?}}
No, all potentially sensitive or confidential information has been removed.
\paragraph{\textbf{Does the dataset contain information that might be considered inappropriate or offensive?}}
No.

%%%%%%%%%%%%%%%%%%%%%%%%%%%%%%%%%%%%%%%%%%%%%%%%%%%%%%%%%%%%%%%%%%%%%%
% ClinDiagnosES
%%%%%%%%%%%%%%%%%%%%%%%%%%%%%%%%%%%%%%%%%%%%%%%%%%%%%%%%%%%%%%%%%%%%%%
\newpage
\section*{ClinDiagnosES}

The ClinDiagnosES \citep{lenguajenaturalai2024medicalexpertes} % (Cite removed for review)
dataset comprises clinical cases accompanied by corresponding diagnoses, collected directly from healthcare professionals across multiple specialties in Europe. It is intended for evaluating LLMs’ diagnostic reasoning abilities.

\subsection*{Motivation for Dataset Creation}
\paragraph{\textbf{Why was the dataset created?}}
ClinDiagnosES was created to assess the ability of LLMs to generate accurate diagnoses based on clinical case descriptions.
\paragraph{\textbf{What (other) tasks could the dataset be used for?}}
Besides diagnostic evaluation, it can be used to study medical reasoning; however, it is not suitable for treatment planning tasks.
\paragraph{\textbf{Who funded the creation of the dataset?}}
The dataset was developed through a collaboration between LenguajeNatural.AI % an NLP startup\footnote{Name removed for review.}
and healthcare professionals.

\subsection*{Dataset Composition}
\paragraph{\textbf{What are the instances?}}
Each instance consists of a clinical case description along with its corresponding diagnosis.
\paragraph{\textbf{Are relationships between instances made explicit in the data?}}
No, there are no explicit relationships between instances.
\paragraph{\textbf{How many instances of each type are there?}}
The dataset contains 62 examples.
\paragraph{\textbf{What data does each instance consist of?}}
Each instance includes text data representing a clinical case and its associated diagnosis.
\paragraph{\textbf{Is everything included or does the data rely on external resources?}}
The dataset is self-contained.
\paragraph{\textbf{Are there recommended data splits or evaluation measures?}}
No specific splits are recommended; it is intended for evaluation purposes.

\subsection*{Data Collection Process}
\paragraph{\textbf{How was the data collected?}}
Data was collected directly from healthcare professionals across various medical specialties.
\paragraph{\textbf{Who was involved in the data collection process?}}
Healthcare professionals from fields such as cardiology, traumatology, emergency medicine, psychiatry, neurology, dermatology, ENT-laryngology, and anesthesia contributed.
\paragraph{\textbf{Over what time-frame was the data collected?}}
The data was collected in 2024.
\paragraph{\textbf{How was the data associated with each instance acquired?}}
Each clinical case was accompanied by a diagnosis provided by a medical expert.
\paragraph{\textbf{Does the dataset contain all possible instances?}}
It is a curated collection and does not encompass every possible clinical case.
\paragraph{\textbf{If the dataset is a sample, then what is the population?}}
The dataset represents a curated sample of clinical cases from European medical centers.
\paragraph{\textbf{Is there information missing from the dataset and why?}}
No, all necessary information is included.
\paragraph{\textbf{Are there any known errors, sources of noise, or redundancies in the data?}}
The dataset has been reviewed to minimize errors and inconsistencies.

\subsection*{Data Preprocessing}
\paragraph{\textbf{What preprocessing/cleaning was done?}}
The clinical cases and diagnoses were formatted using a standardized template with minimal cleaning.
\paragraph{\textbf{Was the “raw” data saved in addition to the preprocessed/clean data?}}
Yes, the raw clinical texts and diagnoses are preserved.
\paragraph{\textbf{Is the preprocessing software available?}}
No specific preprocessing software was utilized.
\paragraph{\textbf{Does this dataset collection/processing procedure achieve the motivation for creating the dataset stated in the first section of this datasheet?}}
Yes, the procedure ensures the dataset is suitable for evaluating diagnostic reasoning in LLMs.

\subsection*{Dataset Distribution}
\paragraph{\textbf{How is the dataset distributed?}}
The dataset is available on HuggingFace\footnote{\url{https://hf.co/datasets/LenguajeNaturalAI/ClinDiagnosES}}.
\paragraph{\textbf{When will the dataset be released/first distributed?}}
It was released in March 2024.
\paragraph{\textbf{What license (if any) is it distributed under?}}
It is distributed under the CC BY-NC-SA 4.0 license.
\paragraph{\textbf{Are there any fees or access/export restrictions?}}
There are no fees or restrictions.

\subsection*{Dataset Maintenance}
\paragraph{\textbf{Who is supporting/hosting/maintaining the dataset?}}
The dataset is hosted on HuggingFace and maintained by the NLP startup.
\paragraph{\textbf{Will the dataset be updated? How often and by whom?}}
No updates are planned at this time.
\paragraph{\textbf{Is there a repository to link to any/all papers/systems that use this dataset?}}
The dataset is available on HuggingFace\footnote{\url{https://hf.co/datasets/LenguajeNaturalAI/ClinDiagnosES}}.

\subsection*{Legal \& Ethical Considerations}
\paragraph{\textbf{If the dataset relates to people, were they informed about the data collection?}}
The clinical cases were provided by healthcare professionals; any personal details have been removed to ensure anonymity. They were anonymized by the healthcare professionals themselves, before transferring the data to the NLP startup.
\paragraph{\textbf{If it relates to other ethically protected subjects, have appropriate obligations been met?}}
Yes, all obligations have been met and ensured in the data collection process.
\paragraph{\textbf{If it relates to people, were there any ethical review applications/reviews/approvals?}}
Yes, healthcare professionals ensured the ethical review was complete.
\paragraph{\textbf{If it relates to people, were they told what the dataset would be used for and did they consent?}}
Yes, patients were told in advance about the objective of data collection and they provided their consent for this use.
\paragraph{\textbf{If it relates to people, could this dataset expose people to harm or legal action?}}
No, as the data is anonymized by the healthcare professionals.
\paragraph{\textbf{If it relates to people, does it unfairly advantage or disadvantage a particular social group?}}
No.
\paragraph{\textbf{If it relates to people, were they provided with privacy guarantees?}}
Yes, all personal information has been removed by the healthcare professionals.
\paragraph{\textbf{Does the dataset comply with the EU General Data Protection Regulation (GDPR)?}}
Yes, it complies with GDPR.
\paragraph{\textbf{Does the dataset contain information that might be considered sensitive or confidential?}}
No, all potentially sensitive or confidential information has been removed.
\paragraph{\textbf{Does the dataset contain information that might be considered inappropriate or offensive?}}
No.

%%%%%%%%%%%%%%%%%%%%%%%%%%%%%%%%%%%%%%%%%%%%%%%%%%%%%%%%%%%%%%%%%%%%%%
% HumorQA
%%%%%%%%%%%%%%%%%%%%%%%%%%%%%%%%%%%%%%%%%%%%%%%%%%%%%%%%%%%%%%%%%%%%%%
\newpage
\section*{HumorQA}

The HumorQA dataset \citep{lenguajenaturalai2024humorqa},
developed collaboratively by LenguajeNatural.AI and Human Profit Consulting, % an NLP startup and a psychology consulting firm\footnote{Names removed for review.},
focuses on humor classification. It consists of jokes paired with labels indicating the joke type: C/E (Comparison/Exaggeration), JP (Wordplay), R3 (Rule of Three) and AI (Animating the Inanimate). The data set is based on a study involving 94 executives and is intended to evaluate the ability of LLMs to understand and classify humor.

\subsection*{Motivation for Dataset Creation}
\paragraph{\textbf{Why was the dataset created?}}
HumorQA was created to assess the ability of LLMs to recognize and classify different types of humor.
\paragraph{\textbf{What (other) tasks could the dataset be used for?}}
It can also be used for research on sentiment analysis and humor recognition, although its primary purpose is humor classification.
\paragraph{\textbf{Who funded the creation of the dataset?}}
The dataset was developed through a collaboration between an NLP startup and a psychology consulting firm.

\subsection*{Dataset Composition}
\paragraph{\textbf{What are the instances?}}
Each instance comprises a joke and its corresponding humor-type label.
\paragraph{\textbf{Are relationships between instances made explicit in the data?}}
No, there are no explicit relationships between instances.
\paragraph{\textbf{How many instances of each type are there?}}
The dataset contains 51 examples.
\paragraph{\textbf{What data does each instance consist of?}}
Each instance includes text data representing a joke and a label indicating its humor category.
\paragraph{\textbf{Is everything included or does the data rely on external resources?}}
The dataset is self-contained.
\paragraph{\textbf{Are there recommended data splits or evaluation measures?}}
No specific splits are recommended; it is intended for evaluation purposes.

\subsection*{Data Collection Process}
\paragraph{\textbf{How was the data collected?}}
Jokes were collected and curated as part of a research study involving humor workshops and interviews with 94 executives.
\paragraph{\textbf{Who was involved in the data collection process?}}
The data collection involved humor experts at Human Profit Consulting along with participating executives.
\paragraph{\textbf{Over what time-frame was the data collected?}}
The data was collected in 2024.
\paragraph{\textbf{How was the data associated with each instance acquired?}}
Jokes were labeled according to a predefined categorization based on the study’s methodology.
\paragraph{\textbf{Does the dataset contain all possible instances?}}
It is a curated sample representing various humor styles.
\paragraph{\textbf{If the dataset is a sample, then what is the population?}}
The sample represents humorous content identified in a study with executives from diverse sectors.
\paragraph{\textbf{Is there information missing from the dataset and why?}}
No, all relevant information is included.
\paragraph{\textbf{Are there any known errors, sources of noise, or redundancies in the data?}}
The dataset has been thoroughly reviewed; no significant errors or redundancies have been identified.

\subsection*{Data Preprocessing}
\paragraph{\textbf{What preprocessing/cleaning was done?}}
The jokes and labels were formatted into a standardized template with minimal preprocessing.
\paragraph{\textbf{Was the “raw” data saved in addition to the preprocessed/clean data?}}
Yes, the original joke texts are preserved.
\paragraph{\textbf{Is the preprocessing software available?}}
No specific preprocessing software was used.
\paragraph{\textbf{Does this dataset collection/processing procedure achieve the motivation for creating the dataset stated in the first section of this datasheet?}}
Yes, the curation process supports the evaluation of humor classification by LLMs.

\subsection*{Dataset Distribution}
\paragraph{\textbf{How is the dataset distributed?}}
The dataset is available on HuggingFace\footnote{\url{https://hf.co/datasets/LenguajeNaturalAI/HumorQA}}.
\paragraph{\textbf{When will the dataset be released/first distributed?}}
It was released in March 2024.
\paragraph{\textbf{What license (if any) is it distributed under?}}
It is distributed under the CC BY-NC-SA 4.0 license.
\paragraph{\textbf{Are there any fees or access/export restrictions?}}
There are no fees or restrictions.

\subsection*{Dataset Maintenance}
\paragraph{\textbf{Who is supporting/hosting/maintaining the dataset?}}
The dataset is hosted on HuggingFace by the NLP startup.
\paragraph{\textbf{Will the dataset be updated? How often and by whom?}}
No updates are planned at this time.
\paragraph{\textbf{Is there a repository to link to any/all papers/systems that use this dataset?}}
The dataset is available on HuggingFace\footnote{\url{https://hf.co/datasets/LenguajeNaturalAI/HumorQA}}.

\subsection*{Legal \& Ethical Considerations}
\paragraph{\textbf{If the dataset relates to people, were they informed about the data collection?}}
The dataset is based on humorous content and research; it does not involve personal data.
\paragraph{\textbf{If it relates to other ethically protected subjects, have appropriate obligations been met?}}
Not applicable.
\paragraph{\textbf{If it relates to people, were there any ethical review applications/reviews/approvals?}}
Not applicable.
\paragraph{\textbf{If it relates to people, were they told what the dataset would be used for and did they consent?}}
Not applicable.
\paragraph{\textbf{If it relates to people, could this dataset expose people to harm or legal action?}}
No.
\paragraph{\textbf{If it relates to people, does it unfairly advantage or disadvantage a particular social group?}}
No.
\paragraph{\textbf{If it relates to people, were they provided with privacy guarantees?}}
Not applicable.
\paragraph{\textbf{Does the dataset comply with the EU General Data Protection Regulation (GDPR)?}}
Yes, it complies with GDPR.
\paragraph{\textbf{Does the dataset contain information that might be considered sensitive or confidential?}}
No.
\paragraph{\textbf{Does the dataset contain information that might be considered inappropriate or offensive?}}
No.

%%%%%%%%%%%%%%%%%%%%%%%%%%%%%%%%%%%%%%%%%%%%%%%%%
\newpage
\section*{RagQuAS}

The Retrieval-Augmented-Generation and Question-Answering in Spanish (RagQuAS) dataset \citep{iic2024ragquas},
created by Instituto de Ingeniería del Conocimiento, % a research institute\footnote{Name removed for review.},
is a high-quality monolingual Spanish dataset designed to evaluate models in retrieval-augmented generation (RAG) and question-answering tasks. 
It consists of 201 examples covering a wide range of knowledge domains, such as hobbies, linguistics, health, astronomy, and customer service.
Each example includes a question, multiple context passages extracted from different documents, and a gold-standard answer.
This dataset is particularly useful for assessing a model's ability to retrieve relevant information from multiple sources and generate accurate, contextually appropriate responses.

\subsection*{Motivation for Dataset Creation }

\paragraph{Why was the dataset created?} (e.g., were there specific tasks in mind, or a specific gap that needed to be filled?) 
RagQuAS was created to provide high-quality examples of questions and answers with related contexts that can be used to evaluate models in retrieval-augmented generation (RAG) and question-answering tasks.

\paragraph{What (other) tasks could the dataset be used for?} Are there obvious tasks for which it should not be used? 
There are no recommended uses for this dataset other than evaluation. 

\paragraph{Who funded the creation of the dataset?} If there is an associated grant, provide the grant number. 
The dataset was created and funded by Instituto de Ingeniería de Conocimiento. 

\subsection*{Dataset Composition} 

\paragraph{What are the instances?} (that is, examples; e.g., documents, images, people, countries) Are there multiple types of instances? (e.g., movies, users, ratings; people, interactions between them; nodes, edges) 
Each instance consists of several categories of text: the topic, a question, an indicator of the variant of the question (this represents questions with linguistic differences but pertaining to the same contexts than other questions), an answer, one to five contexts, one to five complete documents from where the contexts were extracted and the links to these documents. 

\paragraph{Are relationships between instances made explicit in the data} (e.g., social network links, user/movie ratings, etc.)? 
There are no known relationships between instances. 

\paragraph{How many instances of each type are there? }
The dataset consists of 201 examples. 

\paragraph{What data does each instance consist of? }“Raw” data (e.g., unprocessed text or images)? Features/attributes? Is there a label/target associated with instances? If the instances are related to people, are subpopulations identified (e.g., by age, gender, etc.) and what is their distribution? 
The instances consist of text data and are labeled with the corresponding topic. 

\paragraph{Is everything included or does the data rely on external resources?} % (e.g., websites, tweets, datasets) If external resources, a) are there guarantees that they will exist, and remain constant, over time; b) is there an official archival version. Are there licenses, fees or rights associated with any of the data? 
Everything is included in the dataset.   

\paragraph{Are there recommended data splits or evaluation measures?} % (e.g., training, development, testing; accuracy/AUC) 
Since the dataset is intended for testing, there is no recommended split. 

\subsection*{Data Collection Process}

\paragraph{How was the data collected?} %(e.g., hardware apparatus/sensor, manual human curation, software program, software interface/API; how were these constructs/measures/methods validated?) 
The data for the contexts was gathered from different sources manually with the help of generative models (to suggest web searches and results). The rest of the dataset was curated and created manually.  

\paragraph{Who was involved in the data collection process? How were they compensated? }
The data was collected by computational linguists and data scientists from the research institute. 

\paragraph{Over what time-frame was the data collected?} % Does the collection time-frame match the creation time-frame? 
The data was collected during 2023, when the dataset was created. 

\paragraph{How was the data associated with each instance acquired?} % Was the data directly observable (e.g., raw text, movie ratings), reported by subjects (e.g., survey responses), or indirectly inferred/derived from other data (e.g., part of speech tags; model-based guesses for age or language)? If the latter two, were they validated/verified and if so how? 
The question-answer pairs were created and revised by computational linguists. 

\paragraph{Does the dataset contain all possible instances?} % Or is it, for instance, a sample (not necessarily random) from a larger set of instances? 
The dataset is composed of selected instances of a dataset created by the research institute. 

\paragraph{If the dataset is a sample, then what is the population?} % What was the sampling strategy (e.g., deterministic, probabilistic with specific sampling probabilities)? Is the sample representative of the larger set (e.g., geographic coverage)? If not, why not (e.g., to cover a more diverse range of instances)? How does this affect possible uses? 
This dataset is a 24\% sample of the original complete datasets. The instances were randomly selected from the original dataset. 

\paragraph{Is there information missing from the dataset and why?} % (this does not include intentionally dropped instances; it might include, e.g., redacted text, withheld documents) Is this data missing because it was unavailable? 
There is no data missing. 

\paragraph{Are there any known errors, sources of noise, or redundancies in the data?} 
There are no known errors because the revision process ensured the data is as clean and error free as possible. 

\subsection*{Data Preprocessing}

\paragraph{What preprocessing/cleaning was done?} % (e.g., discretization or bucketing, tokenization, part-of-speech tagging, SIFT feature extraction, removal of instances, processing of missing values, etc.) 
The text contained in context and document part of each instance in the dataset has not undergone any preprocessing or changes. The questions were created manually by computational linguists following a rigorous methodology and were subjected to revision afterwards. The answers were carefully curated and revised by linguists from generated texts. 

\paragraph{Was the “raw” data saved in addition to the preprocessed/clean data?} % (e.g., to support unanticipated future uses) 
No, the text in the dataset is the raw data. 

\paragraph{Is the preprocessing software available? }
No preprocessing software was used. 

\paragraph{Does this dataset collection/processing procedure achieve the motivation for creating the dataset stated in the first section of this datasheet? }
Yes, the methodology used when creating the dataset ensures it is sufficiently varied so it can be used to evaluate a model on a wide range of topics. However, there are some potential limitations in the dataset which might slightly bias the data towards particular topics, because not all topics included have the exact same representation in the dataset, and obviously it was not possible to cover all topics in existence.  

\subsection*{Dataset Distribution}

\paragraph{How is the dataset distributed?}% (e.g., website, API, etc.; does the data have a DOI; is it archived redundantly?) 
The dataset is available in HuggingFace\footnote{\url{https://hf.co/datasets/IIC/RagQuAS}}.

\paragraph{When will the dataset be released/first distributed?} % (Is there a canonical paper/reference for this dataset?) 
The dataset was released in 2024. 

\paragraph{What license (if any) is it distributed under?} Are there any copyrights on the data? 
The dataset is licensed under \href{https://creativecommons.org/licenses/by-nc-sa/4.0}{CC BY-NC-SA 4.0}. 

\paragraph{Are there any fees or access/export restrictions? }
There are no fees or restrictions. 

\subsection*{Dataset Maintenance}

\paragraph{Who is supporting/hosting/maintaining the dataset?} How does one contact the owner/curator/manager of the dataset? 
The dataset is hosted at HuggingFace, and the research institute can be contacted through email \href{mailto:contacto.iic@iic.uam.es}{contacto.iic@iic.uam.es}.

\paragraph{Will the dataset be updated? How often and by whom?} How will updates/revisions be documented and communicated? Is there an erratum? 
It is not planned to update the dataset at the moment.   

\paragraph{Is there a repository to link to any/all papers/systems that use this dataset?} 
No.  

\subsection*{Legal \& Ethical Considerations}

\paragraph{If the dataset relates to people or was generated by people, were they informed about the data collection?} % (e.g., datasets that collect writing, photos, interactions, transactions, etc.) 
Not applicable. The data was collected from public web sources, and does not contain sensitive personal information. 

\paragraph{If it relates to other ethically protected subjects, have appropriate obligations been met? } % (e.g., medical data might include information collected from animals) 
Not applicable.    

\paragraph{If it relates to people, were there any ethical review applications/reviews/approvals? } % (e.g. Institutional Review Board applications) 
Not applicable. 

\paragraph{If it relates to people, were they told what the dataset would be used for and did they consent? } %What community norms exist for data collected from human communications? If consent was obtained, how? Were the people provided with any mechanism to revoke their consent in the future or for certain uses? 
Not applicable. 

\paragraph{If it relates to people, could this dataset expose people to harm or legal action?} % (e.g., financial social or otherwise) What was done to mitigate or reduce the potential for harm? 
Not applicable. 

\paragraph{If it relates to people, does it unfairly advantage or disadvantage a particular social group?} % In what ways? How was this mitigated? 
Not applicable. 

\paragraph{If it relates to people, were they provided with privacy guarantees?} % If so, what guarantees and how are these ensured? 
Not applicable. 

\paragraph{Does the dataset comply with the EU General Data Protection Regulation (GDPR)?} % Does it comply with any other standards, such as the US Equal Employment Opportunity Act? 
The dataset complies with GDPR. 

\paragraph{Does the dataset contain information that might be considered sensitive or confidential?} % (e.g., personally identifying information) 
No. 

\paragraph{Does the dataset contain information that might be considered inappropriate or offensive? }
No.

%%%%%%%%%%%%%%%%%%%%%%%%%%%%%%%%%%%%%%%%%%%%%%%%%%%%%%%%%%%%%%%%%%%%%%
% SpaLawEx
%%%%%%%%%%%%%%%%%%%%%%%%%%%%%%%%%%%%%%%%%%%%%%%%%%%%%%%%%%%%%%%%%%%%%%
\newpage
\section*{SpaLawEx}
The SpaLawEx dataset \citep{lenguajenaturalai2024spalawex}
consists of multiple-choice legal questions extracted from Spanish Bar Examination papers of 2022 and 2023. Each instance includes a legal question along with four answer options (A, B, C, and D).

\subsection*{Motivation for Dataset Creation}
\paragraph{\textbf{Why was the dataset created?}}
SpaLawEx was created to evaluate the legal reasoning and knowledge of LLMs within the context of Spanish law using multiple-choice questions.
\paragraph{\textbf{What (other) tasks could the dataset be used for?}}
In addition to benchmarking legal question answering systems, it may be used for legal education; it is not intended for non-legal tasks.
\paragraph{\textbf{Who funded the creation of the dataset?}}
The dataset was developed by an NLP startup,
with contributions from legal experts.

\subsection*{Dataset Composition}
\paragraph{\textbf{What are the instances?}}
Each instance is a multiple-choice legal question accompanied by four answer options.
\paragraph{\textbf{Are relationships between instances made explicit in the data?}}
No, there are no explicit relationships between instances.
\paragraph{\textbf{How many instances of each type are there?}}
The dataset contains 119 examples.
\paragraph{\textbf{What data does each instance consist of?}}
Each instance comprises text data, including a legal question and its four answer options (A, B, C, and D).
\paragraph{\textbf{Is everything included or does the data rely on external resources?}}
The dataset is self-contained, extracted from publicly available examination papers.
\paragraph{\textbf{Are there recommended data splits or evaluation measures?}}
No specific splits are recommended; the dataset is intended for evaluation purposes.

\subsection*{Data Collection Process}
\paragraph{\textbf{How was the data collected?}}
Data were extracted from official Spanish Bar Examination papers from 2022 and 2023.
\paragraph{\textbf{Who was involved in the data collection process?}}
The extraction was performed by the developers at an NLP startup,
with input from legal experts.
\paragraph{\textbf{Over what time-frame was the data collected?}}
The data was collected in 2024.
\paragraph{\textbf{How was the data associated with each instance acquired?}}
Questions and answer options were directly extracted from exam documents.
\paragraph{\textbf{Does the dataset contain all possible instances?}}
It is a comprehensive collection of questions from the specified examination periods. However, it is not exhaustive and it does not contain all possible instances.
\paragraph{\textbf{If the dataset is a sample, then what is the population?}}
It represents the pool of questions from the Spanish Bar Examinations of 2022 and 2023.
\paragraph{\textbf{Is there information missing from the dataset and why?}}
No, all relevant information is included.
\paragraph{\textbf{Are there any known errors, sources of noise, or redundancies in the data?}}
The dataset has been checked for accuracy; any minor extraction errors are not known to be significant.

\subsection*{Data Preprocessing}
\paragraph{\textbf{What preprocessing/cleaning was done?}}
The exam questions and answer options were formatted into a standardized template with minimal cleaning.
\paragraph{\textbf{Was the “raw” data saved in addition to the preprocessed/clean data?}}
Yes, the original extracted text is preserved.
\paragraph{\textbf{Is the preprocessing software available?}}
No specific preprocessing software was used.
\paragraph{\textbf{Does this dataset collection/processing procedure achieve the motivation for creating the dataset stated in the first section of this datasheet?}}
Yes, the process ensures the dataset is suitable for evaluating legal reasoning in LLMs.

\subsection*{Dataset Distribution}
\paragraph{\textbf{How is the dataset distributed?}}
The dataset is available on HuggingFace\footnote{\url{https://hf.co/datasets/LenguajeNaturalAI/SpaLawEx}}.
\paragraph{\textbf{When will the dataset be released/first distributed?}}
It was released in March 2024.
\paragraph{\textbf{What license (if any) is it distributed under?}}
It is distributed under the CC BY-NC-SA 4.0 license.
\paragraph{\textbf{Are there any fees or access/export restrictions?}}
There are no fees or restrictions.

\subsection*{Dataset Maintenance}
\paragraph{\textbf{Who is supporting/hosting/maintaining the dataset?}}
The dataset is hosted on HuggingFace by the NLP startup.
\paragraph{\textbf{Will the dataset be updated? How often and by whom?}}
No updates are planned at this time.
\paragraph{\textbf{Is there a repository to link to any/all papers/systems that use this dataset?}}
No repository has been provided.

\subsection*{Legal \& Ethical Considerations}
\paragraph{\textbf{If the dataset relates to people, were they informed about the data collection?}}
The dataset is derived from public examination materials and does not involve personal data.
\paragraph{\textbf{If it relates to other ethically protected subjects, have appropriate obligations been met?}}
Not applicable.
\paragraph{\textbf{If it relates to people, were there any ethical review applications/reviews/approvals?}}
Not applicable.
\paragraph{\textbf{If it relates to people, were they told what the dataset would be used for and did they consent?}}
Not applicable.
\paragraph{\textbf{If it relates to people, could this dataset expose people to harm or legal action?}}
No.
\paragraph{\textbf{If it relates to people, does it unfairly advantage or disadvantage a particular social group?}}
No.
\paragraph{\textbf{If it relates to people, were they provided with privacy guarantees?}}
Not applicable.
\paragraph{\textbf{Does the dataset comply with the EU General Data Protection Regulation (GDPR)?}}
Yes, it complies with GDPR.
\paragraph{\textbf{Does the dataset contain information that might be considered sensitive or confidential?}}
No.
\paragraph{\textbf{Does the dataset contain information that might be considered inappropriate or offensive?}}
No.

\newpage
\section*{TELEIA}

The TELEIA \citep{spanish_benchmark_teleia}
dataset is intended for the evaluation of Spanish language knowledge focusing on reading comprehension and grammatical competence. The dataset is designed as a set of multiple-choice questions that have the same format and level as those used in several Spanish evaluation tests for humans. The questions are divided into three blocks which resemble existing tests of Spanish for foreign learners and University access. In total, one hundred questions are included that have been prepared and revised by experts on Spanish language, and that have been validated by comparing the results with the original exams.

\subsection*{Motivation for Dataset Creation }

\paragraph{Why was the dataset created?} The main motivation was to have a simple test to evaluate the competence of LLMs in Spanish, similar to tests used with humans.

\paragraph{What (other) tasks could the dataset be used for?} The test also checks reading comprehension and thus can be used to evaluate natural language understanding.

\paragraph{Who funded the creation of the dataset?}
The development of the dataset was supported by the FUN4DATE (PID2022-136684OB-C22) project funded by the Spanish Agencia Estatal de Investigación (AEI) 10.13039/501100011033.

\subsection*{Dataset Composition} 

\paragraph{What are the instances?} The test is made of multiple-choice questions.

\paragraph{Are relationships between instances made explicit in the data} No.

\paragraph{How many instances of each type are there?} The dataset consists of 100 questions. 

\paragraph{What data does each instance consist of? } Each question has a text presenting the question and four answer options, of which only one is correct.

\paragraph{Is everything included or does the data rely on external resources?}  Everything is included in the dataset.

\paragraph{Are there recommended data splits or evaluation measures?} No.

\subsection*{Data Collection Process }

\paragraph{How was the data collected?} Questions were formulated and peer-reviewed by experts in Spanish. 

\paragraph{Who was involved in the data collection process? } Experts in Spanish who participated as researchers in our group.

\paragraph{Over what time-frame was the data collected?} The questions were created during the spring of 2024.

\paragraph{How was the data associated with each instance acquired?} Data was created by experts.

\paragraph{Does the dataset contain all possible instances?} Questions are examples, and many other similar questions can be formulated. 

\paragraph{If the dataset is a sample, then what is the population?} Not applicable.

\paragraph{Is there information missing from the dataset and why?} No. 

\paragraph{Are there any known errors, sources of noise, or redundancies in the data?} No.

\subsection*{Data Preprocessing} 

\paragraph{What preprocessing/cleaning was done?} None. 

\paragraph{Was the “raw” data saved in addition to the preprocessed/clean data?} Not applicable.

\paragraph{Is the preprocessing software available? } Not applicable.

\paragraph{Does this dataset collection/processing procedure achieve the motivation for creating the dataset stated in the first section of this datasheet? } Yes.

\subsection*{Dataset Distribution} 

\paragraph{How is the dataset distributed?} Websites. 

\paragraph{When will the dataset be released/first distributed?} Data is available since July 2024.

\paragraph{What license (if any) is it distributed under?} No license or restrictions are applicable.

\paragraph{Are there any fees or access/export restrictions? } No.

\subsection*{Dataset Maintenance }

\paragraph{Who is supporting/hosting/maintaining the dataset?} The dataset is hosted at Zenodo\footnote{\url{https://zenodo.org/records/12571763}} providing contact details for all the authors.

\paragraph{Will the dataset be updated?} No updates are expected, but the repository supports versioning. 

\paragraph{Is there a repository to link to any/all papers/systems that use this dataset?} No.

\subsection*{Legal \& Ethical Considerations }

\paragraph{If the dataset relates to people (e.g., their attributes) or was generated by people, were they informed about the data collection?} Not applicable.

\paragraph{If it relates to other ethically protected subjects, have appropriate obligations been met? } Not applicable. 

\paragraph{If it relates to people, were there any ethical review applications/reviews/approvals? } Not applicable. 

\paragraph{If it relates to people, were they told what the dataset would be used for and did they consent? } Not applicable. 

\paragraph{If it relates to people, could this dataset expose people to harm or legal action?} Not applicable.  

\paragraph{If it relates to people, does it unfairly advantage or disadvantage a particular social group?} Not applicable. 

\paragraph{If it relates to people, were they provided with privacy guarantees?} Not applicable. 

\paragraph{Does the dataset comply with the EU General Data Protection Regulation (GDPR)?} Yes. 

\paragraph{Does the dataset contain information that might be considered sensitive or confidential?} No.

\paragraph{Does the dataset contain information that might be considered inappropriate or offensive?} No.

\end{document}